\documentclass{article} 
\usepackage{style,times}

\usepackage{amsmath,amsfonts,bm}

\def\eqref#1{equation~\ref{#1}}

\def\1{\bm{1}}

\DeclareMathAlphabet{\mathsfit}{\encodingdefault}{\sfdefault}{m}{sl}
\SetMathAlphabet{\mathsfit}{bold}{\encodingdefault}{\sfdefault}{bx}{n}

\usepackage[utf8]{inputenc} 
\usepackage[T1]{fontenc}    
\usepackage{url}            
\usepackage{booktabs}       
\usepackage{amsfonts}       
\usepackage{nicefrac}       
\usepackage{microtype}      

\usepackage{graphicx}
\usepackage{subcaption}
\usepackage{adjustbox}
\usepackage{booktabs} 
\usepackage{multirow, threeparttable}
\usepackage{wrapfig}
\usepackage{caption}
\usepackage{hhline}
\usepackage{enumitem}
\usepackage{tikz}
\usepackage{tcolorbox}
\usepackage{wrapfig}

\usepackage{titlesec}
\titlespacing*{\section}
{0pt}{4pt}{4pt}

\usepackage{algorithm2e}
\usepackage{bbm}
\RestyleAlgo{ruled}

\usepackage[dvipsnames]{xcolor}
\usepackage{pifont}
\usepackage[unicode=true,psdextra]{hyperref}
\usepackage{cleveref}

\definecolor{lightblue}{HTML}{B0C0FC}
\definecolor{violet_soft}{HTML}{CDA2BE}
\usepackage{tikz}
\DeclareRobustCommand\circled[1]{\tikz[baseline=(char.base)]{\node[shape=circle,draw=blue,minimum size=0.4cm,inner sep=0pt,fill=lightblue] (char) {\fontfamily{phv}\selectfont \scriptsize \textbf{#1}};}}
\newcommand*\circledred[1]{%
\tikz[baseline=(char.base)]{
  \node[shape=circle, draw=BrickRed!60, fill=BrickRed!10, thick, inner sep=1pt] (char) {\scriptsize\textsf{#1}};
}}

\newcommand*\challenges[1]{\tikz[baseline=(char.base)]{\node[shape=circle,draw=black,minimum size=0.35cm,inner sep=0pt,fill=violet_soft] (char) {\fontfamily{phv}\selectfont \scriptsize #1};}}

\newcommand{\greentext}[1]{\textcolor{RedViolet}{#1}}
\newcommand{\bluetext}[1]{\textcolor{NavyBlue}{#1}}

\newcommand{\LabeledDataset}{\bluetext{\mathcal{D}_{\mathrm{GT}}}}
\newcommand{\UnlabeledDataset}{\greentext{\mathcal{D}_{\mathrm{ZCP}}}}
\newcommand{\LabeledPairs}{\bluetext{\mathcal{P}_{\mathrm{GT}}}}
\newcommand{\UnlabeledPairs}{\greentext{\mathcal{P}_{\mathrm{ZCP}}}}

\usepackage{amsmath}
\usepackage{amssymb}
\usepackage{mathtools}
\usepackage{amsthm}
\usepackage{siunitx}
\usepackage{xspace}
\usepackage{tabularx}

\definecolor{darkblue}{HTML}{6C8EBF}

\newtcolorbox{mybox}{colback=white,colframe=darkblue,boxsep=1pt,
    left=2pt,
    right=2pt,
    top=1pt,
    bottom=1pt,before skip=2pt,
    after skip=2pt}

\newtcolorbox{myboxEQ}{colback=white,colframe=darkblue,boxsep=0pt,
    left=1pt,
    right=1pt,
    top=1pt,
    bottom=0pt,
    before skip=2pt,
    after skip=2pt}

\newcommand{\framework}{\textsc{ppNAS}\xspace}

\title{Prediction-powered Neural Architecture Search}

\author{Pascal Janetzky\\
Bosch Center for Artificial Intelligence\\
LMU Munich\\
Munich Center for Machine Learning(MCML) \\
\And
Yuxin Wang\\
LMU Munich\\
Munich Center for Machine Learning (MCML)\\
\AND
Michael Klar\\
Bosch Center for Artificial Intelligence\\
\And
Stefan Feuerriegel\\
LMU Munich\\
Munich Center for Machine Learning (MCML)\\
}

\iclrfinalcopy 
\begin{document}

\maketitle

\begin{abstract}
Evaluating candidate architectures in neural architecture search (NAS) faces an inherent trade-off: on the one hand, reliable performance labels are limited because training and evaluating architectures is costly; on the other hand, zero-cost proxies (ZCPs) are cheap to compute at large scale but can be noisy. Yet, how to effectively combine these two sources of supervision remains unclear. In this paper, we propose \framework, a novel prediction-powered inference (PPI) approach for NAS. \framework fuses (1) a small set of architectures with observed performance labels and (2) a large set of architectures with ZCP information. To combine these two sources of supervision, \framework exploits the ordinal information provided by ZCPs to construct additional pairwise ranking supervision, while PPI debiases systematic discrepancies between ZCP-based and true performance rankings. We evaluate \framework in end-to-end predictor-based NAS, where it achieves state-of-the-art under limited evaluation budgets. To the best of our knowledge, \framework is the first prediction-powered approach for label-efficient NAS.

\end{abstract}

\section{Introduction}
The goal of neural architecture search (NAS) is to identify high-performing neural network architectures. NAS has been successfully applied across a wide range of tasks, including image classification \citep{zoph2016neural,real2019regularized}, object detection and semantic segmentation \citep{poyser2024nassurvey}, and natural language processing \citep{elsken2019survey,white2023neural}. However, obtaining reliable performance estimates for candidate architectures typically requires frequent training and evaluation, which is a key bottleneck in NAS, and subject to an inherent trade-off:

\circled{1} ~\emph{Ground-truth performance labels}, on the one hand, can provide reliable supervision for learning predictors, which can estimate the performance of unseen architectures and guide the search toward promising candidate architectures \citep{white2020study,white2023neural,kadlecova2024surprisingly}. However, training such predictors requires labeled data about architectures with corresponding performance estimates, but obtaining these is costly \citep{zoph2016neural,real2019regularized}. As a result, the number of ground-truth labels available for learning accurate performance predictors is limited.

\circled{2}~\emph{Zero-cost proxies (ZCPs)}, on the other hand, can provide cheap signals about the performance of candidate architectures \emph{without} the cost of training \citep{mellor2021neural,abdelfattah2021zero,Krishnakumar2022zerocost}. To do so, ZCPs assign heuristic scores to candidate architectures based on properties of the neural architecture (e.g., summed importance of specific neurons) or based on other derived quantities computed using a small amount of data \citep{abdelfattah2021zero,ning2021evaluating,tanaka2020pruning,mellor2021neural,wang2020picking,lee2018snip}. Prior work suggests that the ordering induced by ZCPs can provide useful ordinal information for ranking different architectures \citep{dudziak2020brp,abdelfattah2021zero,shen2023proxybo}. However, ZCP scores can be noisy and may not reliably reflect true architecture performance.

Together, both ground-truth performance estimates and ZCPs have different strengths and weaknesses for evaluating candidate architectures in terms of quantity, computational cost, and reliability. This motivates our research question: \textbf{\emph{how can we combine supervision signals from} \circled{1} \emph{reliable but scarce ground-truth performance estimates and}  \circled{2} \emph{scalable but noisy ZCPs?}}   

To address this challenge, \emph{we propose \framework, a prediction-powered inference (PPI) approach for label-efficient NAS}. \framework fuses (i)~a small set of ground-truth labeled architectures with (ii)~a substantially larger set of ZCP-only samples. We use both sources to construct pairwise ranking supervision, where the ground-truth performance labels provide reliable rankings, while ZCPs provide abundant but noisy rankings at low cost. To combine these two sources of supervision, \framework exploits the ordinal information provided by ZCPs, while PPI \citep{Angelopoulos2023ppi} debiases systematic discrepancies between the ZCP-based rankings and the ground-truth performance rankings. As a result, this allows \framework to learn an architecture scoring function that can rank unseen candidate architectures for NAS using only a small number of expensive evaluations. We evaluate \framework as a performance-predictor within end-to-end NAS. Our results show that \framework improves the performance prediction of candidate architectures, which leads to large performance gains in end-to-end NAS under limited budgets.

Methodologically, we adapt principles from PPI \citep{Angelopoulos2023ppi,angelopoulos2024ppiefficientpredictionpoweredinference}  for NAS. PPI provides a principled way to combine large-scale, potentially noisy predictions with scarce ground-truth labels to correct systematic prediction errors and enable statistically valid inference \citep{Angelopoulos2023ppi,Zrnic2024crossppi}. To the best of our knowledge, we are the first to leverage PPI for label-efficient NAS. However, our setting offers three challenges that require careful adaptation. \textbf{Challenge} \challenges{a}:~Standard PPI settings focus on estimating a target population parameter. In contrast, in NAS we want to learn a feature-dependent scoring function $g_\theta(X)$ that must generalize to unseen architecture specifications, where the pool of potential architectures is not fixed. This necessitates an adapted prediction-powered objective for ranking, that can then guide architecture search. \textbf{Challenge} \challenges{b}:~PPI requires prediction in the same target space, but ZCPs instead provide proxy-specific scores that can only be interpreted through their relative ordering. We thus need to formulate PPI over pairwise ranking losses rather than pointwise prediction losses. \textbf{Challenge} \challenges{c}:~The resulting pairwise observations are statistically dependent because multiple pairs can share the same architecture. Consequently, standard variance formulas from PPI++ and power-tuning formulas \citep{angelopoulos2024ppiefficientpredictionpoweredinference} cannot be applied directly at the pair level by treating the pairwise observations as independent; here we instead account for the dependence by computing architecture-level Hájek contributions \citep{55313c38-914c-3260-914e-410e7a9dbf9d,doi:https://doi.org/10.1002/9780470316481.ch5} when controlling the contribution of ZCP supervision and estimation uncertainty.

Our \textbf{main contributions} are three-fold:
\textbf{(1)}~We leverage prediction-powered inference for label-efficient NAS. \textbf{(2)}~We derive a tailored PPI framework for this purpose called \framework by developing a rectified pairwise-ranking objective that accounts for systematic discrepancies between ZCP-induced and ground-truth rankings while adaptively weighting the ZCP supervision. \textbf{(3)}~We demonstrate the practical benefits of \framework in end-to-end NAS, where it improves search performance under limited ground-truth evaluation budgets across multiple benchmarks and ZCPs.

\section{Related Work}
We give an overview of three literature streams relevant to our work: (1)~performance prediction in NAS, (2) zero-cost proxies, and (3) prediction-powered inference.

\textbf{Performance prediction in NAS:} Various search strategies have been proposed for NAS, such as reinforcement learning and evolutionary approaches \citep{zoph2016neural,real2019regularized}, weight-sharing methods \citep{bender2018understanding,pham2018efficient,liu2018darts}, and predictor-based NAS \citep{white2020study,white2021bananas,kadlecova2024surprisingly}. Predictor-based methods, which are the focus of our work, learn a machine learning model to predict architecture performance from the overall architecture specification, and then use these predictions to select promising candidate architectures for subsequent expensive evaluation. Prominent examples of predictor-based methods are \textbf{BANANAS} \citep{white2021bananas}, and \textbf{BOHAMIANN} \citep{springenberg2016bayesian}, which directly predict validation accuracy. An alternative is learning the relative architecture performance through ranking \citep{ji2025loss}. Examples of the latter are \textbf{RankNAS} \citep{hu2021ranknas}, \textbf{ProxyBO} \citep{shen2023proxybo}, and \textbf{ReNAS} \citep{xu2021renas}. However, such performance predictors are inherently \emph{supervised} and still rely on costly ground-truth performance evaluations for training; this makes their performance sensitive to the available budget of labeled data. Our approach is \underline{orthogonal} to predictor-based NAS: \framework does not focus on the performance predictor itself but how to leverage the performance predictions for data fusion.

\textbf{Zero-cost proxies:} ZCPs provide noisy but cheap performance estimates, which do \emph{not} require conventional architecture training \citep{abdelfattah2021zero,mellor2021neural,Krishnakumar2022zerocost}. Common examples are \texttt{synflow} \citep{abdelfattah2021zero,tanaka2020pruning} and \texttt{grasp} \citep{wang2020picking}. These ZCPs are then used either standalone \citep{abdelfattah2021zero,mellor2021neural,lin2021zen,li2023zico} or as additional features into supervised performance predictors \citep{white2021bananas,Krishnakumar2022zerocost,white2023neural,kadlecova2024surprisingly}. Other approaches, such as \textbf{ProxyBO} \citep{shen2023proxybo} or \textbf{ZCPBP Warmup} \citep{abdelfattah2021zero}, create ZCP-based rankings to augment the training of a predictor. However, existing approaches from this stream primarily use ZCPs as standalone signals, as additional predictor inputs, or directly within the search procedure. In contrast, a principled way to combine the ZCPs with ground-truth architecture evaluations is missing. Our approach is thus different by using ZCP-induced rankings as \emph{additional} supervision for ground-truth rankings.

\textbf{Prediction-powered inference:} PPI provides a principled framework for rigorous statistical inference by ``fusing'' a small amount of ground-truth labeled data with a much larger set of potentially noisy predictions to improve statistical efficiency while using the ground-truth labels to correct for errors in the predictions \citep{Angelopoulos2023ppi}; \textbf{a brief overview is in \Cref{sec:ppi_recap}}. Several extensions have been developed over the past years. For example, PPI++ extends this approach by a weighting parameter $\lambda$ to improve the statistical efficiency of the resulting estimator \citep{angelopoulos2024ppiefficientpredictionpoweredinference}. Even other works have adapted PPI to specific tasks such as LLM leaderboards \citet{Chatzi2024llmranking,Boyeau2025autoeval,frauen2026nonparametric}. However, existing PPI approaches differ from our setting in several important ways: \textbf{(1)}~In terms of task, PPI has been applied to a range of \emph{inference} problems, but \textbf{not} to label-efficient NAS.\textbf{ (2)}~In terms of estimand, rather than ranking a fixed set of observed items, we must learn a feature-dependent scoring function $g_\theta(X)$ that generalizes to previously unseen architectures ($\rightarrow$ Challenge \challenges{a}). \textbf{(3)}~In terms of method, this requires adapting PPI to the dependence structure induced by pairwise architecture comparisons ($\rightarrow$ Challenge \challenges{c}).

\textbf{Research gap:} To the best of our knowledge, a principled framework for combining (a)~scarce ground-truth architecture evaluations with (b)~large-scale ZCPs is missing. Here, we thus introduce the first PPI-based approach to label-efficient NAS.

\begin{figure}[t]
    \centering
    \includegraphics[width=0.8\linewidth]{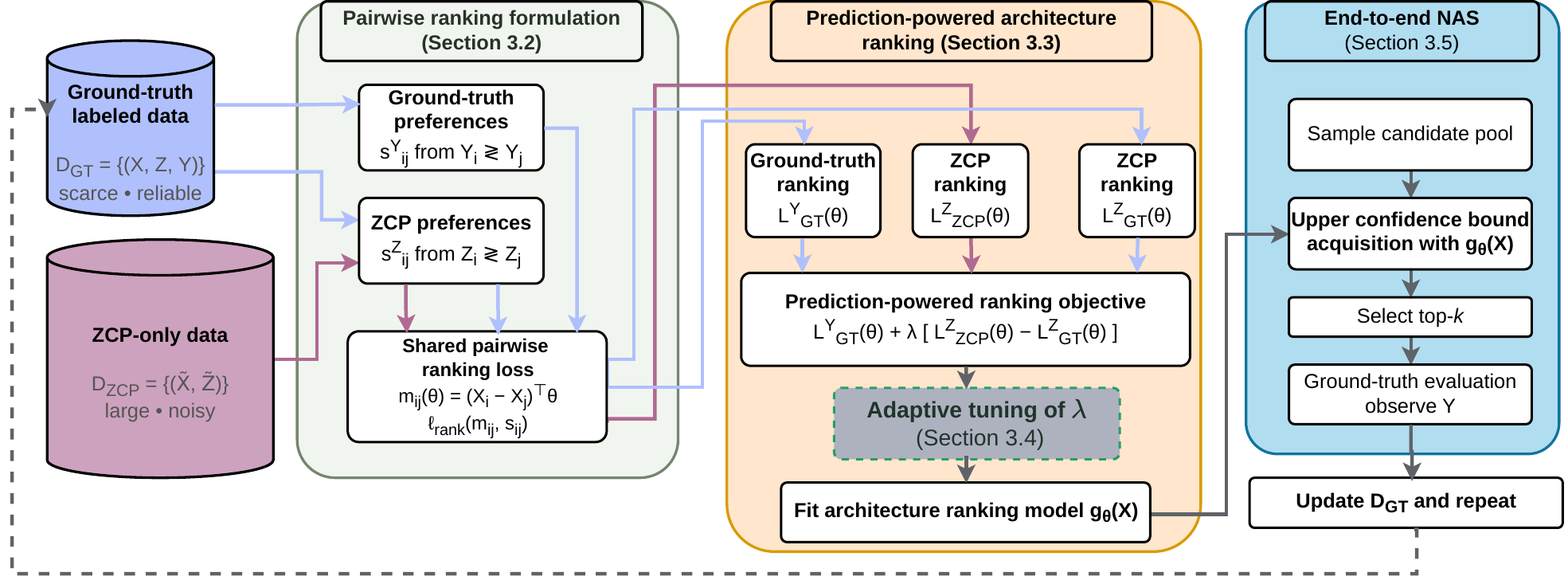}
    \vspace{-3mm}
    \caption{Prediction-powered neural architecture search (\framework). We have access to reliable labeled and noisy zero-cost only datasets (left). From both datasets, we construct pairwise ranking supervision (middle-left). The ranking supervision is then used to learn an architecture ranking model with parameters $\theta$, where the contribution of the noisy data is adaptively weighted (middle-right). The trained model is then used in end-to-end NAS to rank candidate architectures (right).}
    \label{fig:figure_1}
    \vspace{-0.8cm}
\end{figure}

\section{Prediction-powered neural architecture search}
\label{sec:method}

\subsection{Problem Setting}
\label{sec:problem_setting}

\textbf{NAS setup:} We consider a standard NAS setting~\citep{kadlecova2024surprisingly,shen2023proxybo,hu2021ranknas}, where the search space is given by $\mathcal{A}$. Each architecture is represented by features $X \in \mathcal{X}$ and has a vector of ZCP scores $Z \in \mathbb{R}^K$ (e.g., the gradient norm \citep{abdelfattah2021zero}). In addition, each architecture has a ground-truth performance $Y \in \mathbb{R}$, such as validation accuracy after training and evaluating the architecture, but which is costly to obtain.

\textbf{Input:} We have access to two datasets: \textbf{(a)}~a \emph{small labeled} dataset $\LabeledDataset = \{(X_i,Z_i,Y_i)\}_{i=1}^n \overset{\mathrm{iid}}{\sim} \mathbb{P}_{XZY}$, which contains architecture features, ZCP scores, and observed ground-truth (GT) performance; and \textbf{(b)}~a substantially \emph{larger unlabeled} dataset: $\UnlabeledDataset = \{(\widetilde{X}_i,\widetilde{Z}_i)\}_{i=1}^N \overset{\mathrm{iid}}{\sim} \mathbb{P}_{XZ}$, with $n \ll N$. The latter contains architecture features and ZCP scores only, but no ground-truth performance labels. We assume that $\LabeledDataset$ and $\UnlabeledDataset$ are independent and that their architecture--proxy pairs are drawn from the same marginal distribution $\mathbb{P}_{XZ}$.

\begin{mybox}
Our goal is to learn an architecture scoring function $g_\theta:\mathcal{X}\rightarrow\mathbb{R}$ that ranks unseen candidate architectures according to their ground-truth performance. To learn $g_\theta$, we use both (a)~the scarce ground-truth supervision from $\LabeledDataset$ and (b)~the large-scale ZCP information from $\UnlabeledDataset$.
\end{mybox}

Our above task formulation has two important characteristics which will become relevant for our method. \emph{First}, the scoring function $g_\theta$ takes architecture features $X$ as input; this allows the learned architecture scoring function to generalize beyond the architectures contained in $\LabeledDataset$ and $\UnlabeledDataset$ and, therefore, to score previously unseen candidate architectures during NAS based on the given features ($\rightarrow$ Challenge \challenges{a}). \emph{Second}, we note that downstream NAS only requires the relative ordering of architectures; hence, we formulate the learning problem as a ranking task. This also addresses the issue that ZCP scores and ground-truth performance estimates naturally lie on different numerical scales (i.e., ZCPs provide proxy-specific scores, whereas $Y$ may, for example, represent validation accuracy; $\rightarrow$ Challenge \challenges{b})).

\textbf{Overview:} Our \framework has four main components: $\bullet$\,\textbf{Pairwise ranking formulation}. In \Cref{sec:ranking_supervision}, we first define the architecture scoring function and introduce a ranking loss to learn pairwise preferences that is applicable to both ground-truth performance and ZCP scores. $\bullet$\,\textbf{Prediction-powered architecture ranking}. In \Cref{sec:rankppi}, we then combine these two sources of ranking information within a prediction-powered objective.
$\bullet$\,\textbf{Adaptive tuning}. In \Cref{sec:ranking_lambda}, we introduce a parameter $\lambda$ to weight the contribution of the ZCP supervision. $\bullet$\,\textbf{End-to-end NAS with \framework}. In \Cref{sec:rankppi_nas}, we describe how we use \framework in end-to-end search.

\subsection{Pairwise Ranking Formulation}
\label{sec:ranking_supervision}


\textbf{Architecture scoring function.} We learn a \emph{architecture scoring function} $g_\theta(X_i)=X_i^\top\theta$ with parameters $\theta\in\mathbb{R}^d$.\footnote{Note that an intercept cancels out in the pairwise margin later, and we thus omit it here.} For a pair of architectures $i,j$, the predicted ranking margin is $m_{ij}(\theta)=g_\theta(X_i)-g_\theta(X_j)=(X_i-X_j)^\top\theta$. A positive margin indicates that architecture $i$ is predicted to outperform architecture $j$, while a negative margin indicates the opposite. Next, we define the pairwise preference targets that encode the desired ordering of an architecture pair based on either ground-truth performance (in $\LabeledDataset$) or ZCP scores (in both $\LabeledDataset$ and $\UnlabeledDataset$). Importantly, expressing both sources through the same pairwise preference representation allows us to formulate a joint ranking loss across the labeled and unlabeled datasets.

\textbf{Ground-truth pairwise preferences} (for $\LabeledDataset$). For two architectures $i,j \in S_n$ with observed ground-truth performances $Y_i, Y_j$, we define a pairwise preference target via
\begin{equation}
    \label{eq:gold_preference}
    s^Y_{ij}
    =
    \begin{cases}
        1, & Y_i > Y_j,\\
        0, & Y_i < Y_j,\\
        \frac{1}{2}, & Y_i = Y_j,  \end{cases}
\end{equation}
where $s^Y_{ij}=1$ indicates that architecture $i$ should outrank architecture $j$, $s^Y_{ij}=0$ indicates the opposite, and $\frac{1}{2}$ represents a tie.

\textbf{ZCP-induced pairwise preferences} (for $\LabeledDataset$ \emph{and} $\UnlabeledDataset$). Similarly, for a ZCP $Z^{(k)}$, we define ZCP-induced preferences $s^{Z,(k)}_{ij}$ from the ordering of the ZCP scores, analogously to \Cref{eq:gold_preference}:
\begin{equation}
    \label{eq:zcp_preference}
    s^{Z,(k)}_{ij} =
    \begin{cases}
        1, & Z_i^{(k)} > Z_j^{(k)},\\
        0, & Z_i^{(k)} < Z_j^{(k)},\\
        \frac{1}{2}, & Z_i^{(k)} = Z_j^{(k)},
    \end{cases}
\end{equation}
where we assume that larger ZCP scores indicate better performance (this is w.l.o.g. the general interpretation for ZCPs; see \citep{abdelfattah2021zero,Krishnakumar2022zerocost}).

The ranking formulation has a clear benefit: the targets depend \emph{only} on the ordinal information of the ZCP; hence, we do not need to learn a supervised mapping from ZCP scores to validation accuracy. Importantly, both the ground-truth and ZCP-induced preferences are expressed in the same pairwise preference space, which later allows us to adapt PPI ($\rightarrow$ Challenge \challenges{b}).

\textbf{Pairwise ranking loss:} To learn the scoring function from pairwise preferences, we employ a pairwise logistic ranking loss
\begin{equation}
    \label{eq:ranking_loss}
    \ell_{\mathrm{rank}}(m,s)
    =
    -s\log\sigma(m)
    -(1-s)\log\left(1-\sigma(m)\right),
\end{equation}
where $m$ denotes the predicted ranking margin, $s\in[0,1]$ the pairwise preference target, and $\sigma(m)=(1+\exp(-m))^{-1}$ the sigmoid function. Since both ground-truth and ZCP-induced preferences are represented in the same target space, the same ranking loss can be applied to either source of pairwise information.

\subsection{Prediction-Powered Architecture Ranking}
\label{sec:rankppi}


\textbf{Rationale for prediction-powered ranking.}
We now combine the two sources of ranking supervision from $\LabeledDataset$ and $\UnlabeledDataset$ within a prediction-powered objective. The key challenge in applying PPI to our setting is that ZCPs provide proxy-specific scores, while $Y$ denotes architecture performance estimated on a validation set (i.e., accuracy), hence they are on different scales ($\rightarrow$ Challenge \challenges{b}). We therefore apply the PPI principle in the shared space of \emph{pairwise preferences}. To do so, we define three empirical ranking losses: one based on ground-truth preferences constructed from the labeled set, and two based on ZCP-induced preferences constructed from the labeled and unlabeled sets, respectively.

\textbf{Empirical ranking losses.} We define three ranking losses corresponding to the supervision available in $\LabeledDataset$ and $\UnlabeledDataset$: \textbf{(i)}~a ground-truth ranking loss on the labeled set $\LabeledDataset$, \textbf{(ii)}~a ZCP-induced ranking loss on $\LabeledDataset$, and \textbf{(iii)}~a ZCP-induced ranking loss on the unlabeled set $\UnlabeledDataset$. For this, let $\LabeledPairs=\{(i,j):1\leq i<j\leq n\}$ and $\UnlabeledPairs=\{(i,j):1\leq i<j\leq N\}$ denote the sets of unique architecture pairs in $\LabeledDataset$ and $\UnlabeledDataset$, respectively. 

$\bullet$\,\textbf{(i)}~On the labeled set, we define the \textit{empirical ground-truth ranking loss}:
\begin{equation}
    \label{eq:gold_ranking_loss}
    L_\mathrm{GT}^Y(\theta)
    =
    \frac{1}{|\LabeledPairs|}
    \sum_{(i,j)\in\LabeledPairs}
    \ell_{\mathrm{rank}}
    \left(
        m_{ij}(\theta),s^Y_{ij}
    \right).
\end{equation}
This loss measures how well the architecture scoring function agrees with the ordering induced by the observed ground-truth performances. $\bullet$\,\textbf{(ii)}~The \emph{ZCP-induced ranking loss on the \underline{labeled} set} is 
\begin{equation}
    \label{eq:gold_ranking_loss_with_zcp}
    L_\mathrm{GT}^Z(\theta)
    =
    \frac{1}{|\LabeledPairs|}
    \sum_{(i,j)\in\LabeledPairs}
    \ell_{\mathrm{rank}}
    \left(
        m_{ij}(\theta), s^{Z,(k)}_{ij}
    \right).
\end{equation}
which measures agreement with the ZCP-induced preferences available in $\LabeledDataset$. $\bullet$\,\textbf{(iii)}~The \emph{ZCP-induced ranking loss on the \underline{unlabeled} set} is
\begin{equation}
    \label{eq:zcp_ranking loss}
    L_\mathrm{ZCP}^Z(\theta)
    =
    \frac{1}{|\UnlabeledPairs|}
    \sum_{(i,j)\in\UnlabeledPairs}
    \ell_{\mathrm{rank}}
    \left(
        m_{ij}(\theta), s^{Z,(k)}_{ij}
    \right),
\end{equation}
which analogously measures agreement with the ZCP-induced preferences in $\UnlabeledDataset$.

\textbf{Prediction-powered ranking objective.} We adapt the PPI objective from \Cref{sec:ppi_recap} to our ranking setting by replacing its pointwise losses with the three ranking losses defined above. In addition, following PPI++ \citep{angelopoulos2024ppiefficientpredictionpoweredinference}, we introduce a weighting parameter $\lambda\in[0,1]$ to control the contribution of the ZCP-induced supervision.
\begin{myboxEQ}
\begin{equation}
    \label{eq:rankppi_objective}
    \text{The \framework loss is thus }\quad
    \widehat{\theta}_{\lambda}
    =
    \arg\min_{\theta}
    \left\{
        L_\mathrm{GT}^Y(\theta)
        +
        \lambda
        \left[
            L_\mathrm{ZCP}^Z(\theta)
            -
           L_\mathrm{GT}^Z(\theta)
        \right]
    \right\}.
\end{equation}
\end{myboxEQ}
The objective preserves the basic PPI structure while operating entirely on pairwise rankings. The term $L_\mathrm{ZCP}^Z(\theta)$ incorporates the large-scale ZCP-induced supervision from $\UnlabeledDataset$, whereas $L_\mathrm{GT}^Y(\theta)$ and $L_\mathrm{GT}^Z(\theta)$ compare ground-truth and ZCP-induced ranking supervision on the labeled set $\LabeledDataset$. The subtraction of $L_\mathrm{GT}^Z(\theta)$ therefore corrects for systematic discrepancies between the ZCP-induced and ground-truth rankings. Importantly, this correction is performed in the \emph{pairwise preference space}, rather than on the numerical ZCP and performance values themselves. 

The tuning parameter $\lambda$ controls the rectifier, that is, how strongly the ZCP-based supervision contributes, and allows us to reduce its influence when the ZCP-induced rankings are less informative. Hence, setting $\lambda=0$ recovers the ranking objective based on ground-truth supervision, whereas $\lambda=1$ yields the fully rectified prediction-powered objective. Following PPI++ \citep{angelopoulos2024ppiefficientpredictionpoweredinference}, we adaptively select $\lambda$ to improve the statistical efficiency; see next section.

\vspace{-2mm}
\subsection{Adaptive Tuning of $\lambda$}
\label{sec:ranking_lambda}

\textbf{Why standard PPI++ cannot be used.} Standard PPI++ selects $\lambda$ to  minimize the variance of the estimator \citep{angelopoulos2024ppiefficientpredictionpoweredinference}. However, in our setting, the ranking losses are computed over all architecture pairs, and these pairwise observations are \emph{not} independent, as two pairs can share the same architecture (e.g., $(i,j)$ and $(j,k)$ sharing $j$; $\rightarrow$ Challenge \challenges{c}). Hence, treating the constructed pairs as independent observations would underestimate the variance due to sampling dependence. To address this, we propose to characterize variability at the level of the independently sampled architectures rather than at the level of the constructed pairs.

\textbf{Architecture-level variance estimates.} To obtain architecture-level variance estimates, we estimate the first-order Hájek projection \citep{55313c38-914c-3260-914e-410e7a9dbf9d,doi:https://doi.org/10.1002/9780470316481.ch5} of the gradient of the complete-pair ranking loss. For an architecture pair $(i,j)$, the gradient of the pairwise logistic loss is
\begin{equation}
    \label{eq:pairwise_gradient}
    \psi_{ij}(\theta)
    =
    (X_i-X_j)
    \left[
        \sigma\left(m_{ij}(\theta)\right)
        -
        s_{ij}
    \right].
\end{equation}
For a set $B$ of $b$ architectures with generic empirical ranking loss $U_b(\theta)$, we estimate the architecture-level contribution for architecture $i\in B$ by
\begin{equation}
    \label{eq:hajek_influence}
    \widehat{\phi}_i
    =
    2 \Big[
        \cfrac{1}{b-1}
        \sum_{\substack{j\in B, j\neq i}}
        \psi_{ij}(\theta)
        -
        \nabla_\theta U_b(\theta)
    \Big],
\end{equation}
where $\psi_{ij}(\theta)$ uses the corresponding ground-truth or ZCP-induced preference target. Intuitively, $\widehat{\phi}_i$ summarizes the first-order contribution of architecture $i$ to the gradient of the complete-pair ranking loss. Mathematically, this yields one contribution per independently sampled architecture, rather than treating the constructed pairs as independent observations. We compute these contributions separately for the ground-truth ranking loss, the ZCP-ranking loss on the labeled set, and the proxy-ranking loss on the unlabeled set, denoted by $\phi_Y$, $\phi_{Z,n}$, and $\phi_{Z,N}$, respectively.

\textbf{Variance-adaptive weighting of ZCP supervision.} 
We next use these architecture-level H\'ajek contributions to determine how strongly the ZCP-based ranking supervision should enter the rectified objective. On the $\sqrt{n}$ scale, the leading covariance of the rectified estimating equation is

\begin{equation}
    \label{eq:rankppi_gradient_covariance}
    \Omega_\lambda
    =
    \operatorname{Var}
    \left(
        \phi_Y-\lambda\phi_{Z,n}
    \right)
    +
    \frac{n}{N}\lambda^2
    \operatorname{Var}
    \left(
        \phi_{Z,N}
    \right).
\end{equation}
The first term captures the variability of the rectified ranking gradients on the labeled architectures; hence, increasing $\lambda$ can reduce this variability when the proxy-induced gradients are informative about the ground-truth gradients. The second term accounts for the additional sampling variability contributed by the ZCP-induced ranking gradients on the unlabeled architectures.

\textbf{Choice of $\lambda$ in \framework.} We select $\lambda$ by first minimizing the trace of the estimated parameter covariance with fixed $\lambda=1$:
\begin{equation}
    \label{eq:rankppi_lambda_optimization}
    \widehat{\lambda}
    =
    \underset{\lambda\in[0,1]}{\arg\min}
    \;
    \operatorname{tr}
    \left(
        H^\dagger
        \widehat{\Omega}_{\lambda}
        H^{\dagger\top}
    \right),
\end{equation}
where $H^\dagger$ denotes the inverse of the Hessian.\footnote{For numerical stability, we used the pseudoinverse.} 

To compute the Hessian, we proceed as follows: We first fit the  ranking objective using a fixed value of $\lambda=1$ and evaluate the Hessian at the resulting parameter estimate. At the same estimate, we compute the architecture-level H\'ajek contribution covariances and use them to select $\widehat{\lambda}$ according to \Cref{eq:rankppi_lambda_optimization}. We then refit the ranking objective using the selected $\widehat{\lambda}$. The procedure thus provides an adaptive variance reduction similar to PPI++ while accounting for the dependence induced by the pairwise ranking construction.

\subsection{End-to-end NAS with \framework}
\label{sec:rankppi_nas}

\textbf{End-to-end NAS loop.} We now build an end-to-end NAS procedure using \framework to predict the relative performance of candidate architectures and thus guide the search. The procedure proceeds iteratively over iterations $t=1,\ldots$ and consists of two steps. \textbf{(1)~Model update:} At iteration $t$, we fit \framework on the set of architectures evaluated so far, denoted by $\LabeledDataset^{(t)}$, and the unlabeled architectures, denoted by $\UnlabeledDataset^{(t)}$. We then obtain the fitted architecture scoring function $g_{\widehat{\theta}^{(t)}{\widehat{\lambda}}}$ together with the estimated covariance (i.e., $\widehat{\theta}_{\widehat{\lambda}}^{(t)}$). \textbf{(2) Architecture selection:} We use the fitted architecture scoring function to rank the candidate architectures, then select architectures for the next expensive evaluation, and update $\LabeledDataset^{(t)}$ with the new observed ground-truth performances to obtain $\LabeledDataset^{(t+1)}$. We next describe the acquisition rule used for this selection in step~(2).

\textbf{Uncertainty-aware architecture selection.}
Inspired by \citep{Xue2021ucbenas}, we employ an upper confidence bound (UCB) acquisition rule to balance exploitation and exploration. Our acquisition rule is based on the estimated ranking margin and its standard error. We drop $t$ for ease of notation. Let $X_b$ denote the currently best evaluated architecture and $X_{\mathrm{cand}}$ a candidate architecture. With $d=X_{\mathrm{cand}}-X_b$, the predicted ranking margin of $X_{\mathrm{cand}}$ relative to $X_b$ is $d^\top\hat{\theta}_{\lambda}$, with estimated standard error $\sqrt{d^\top\widehat{\mathrm{Cov}}(\hat{\theta}_{\lambda})d}$. We define the acquisition score as
\begin{equation}
    \mathrm{UCB}(X_{\mathrm{cand}};X_b)
    =
    d^\top\hat{\theta}_{\lambda}
    +
    \beta\sqrt{
        d^\top\widehat{\mathrm{Cov}}
        (\hat{\theta}_{\lambda})d
    },
\end{equation}
where $\beta\geq 0$ controls the exploration--exploitation trade-off (we use $\beta=0.5$ in our experiments). The first term favors candidates estimated to outperform the current incumbent, whereas the second favors candidates whose relative ranking is estimated less precisely. At iteration $t$, we evaluate the candidates with the highest acquisition scores and add their observed ground-truth performances to $\LabeledDataset^{(t)}$ to obtain $\LabeledDataset^{(t+1)}$. 

The above procedure has two practical benefits. First, because the acquisition rule operates directly on pairwise ranking margins, it does not require the learned architecture scores to be calibrated to validation accuracy. Second, the covariance estimated by \framework directly provides the uncertainty required for acquisition, allowing uncertainty-aware search with a single fitted ranking model rather than an ensemble.

\vspace{-2mm}
\section{Experimental Setup}
\vspace{-2mm}

\textbf{Benchmark datasets.}
We use NasBench201 (NB201) \citep{dong2020bench}, a tabular benchmark that contains pre-computed validation accuracies and ZCPs for neural architectures trained on \textbf{CIFAR10} \citep{Krizhevsky09learningmultiple}, \textbf{CIFAR100} \citep{Krizhevsky09learningmultiple}, and \textbf{ImageNet16-120} \citep{chrabaszcz2017downsampled,dong2021nats}. To assess whether our findings generalize to other search spaces, we additionally use NASBench101 (NB101) \citep{ying2019bench} and TransNASBench (TNB) \citep{duan2021transnas}.

\textbf{Baselines.}
We compare against the following baselines: \circledred{1}\,\underline{simple baselines}: \textbf{ZCP only} \citep[e.g.][]{Krishnakumar2022zerocost,kadlecova2024surprisingly}, 
\textbf{random search}; \circledred{2}\,\underline{non-ranking baselines}: \textbf{BANANAS} \citep{white2021bananas}, \textbf{BOHAMIANN} \citep{springenberg2016bayesian}, \textbf{SemiNAS} \citep{luo2020semi}; \circledred{3}\,\underline{ranking baselines}: \textbf{ReNAS} \citep{xu2021renas}, \textbf{PWLNAS} \citep{ji2025loss}, \textbf{RankNAS} \citep{hu2021ranknas}, \textbf{ProxyBO} \citep{shen2023proxybo}, and \textbf{ZCPBP Warmup }\citep{abdelfattah2021zero}.

\textbf{ZCPs.}
We focus on the most-widely used ZCPs from the literature: \texttt{synflow} \citep{abdelfattah2021zero,tanaka2020pruning}, \texttt{jacov} and \texttt{nwot} \citep{mellor2021neural}, \texttt{grad\_norm} \citep{abdelfattah2021zero}, \texttt{snip} \citep{lee2018snip}, \texttt{grasp} \citep{wang2020picking}. We encode architectures via a one-hot vector, which is widely used \citep[e.g.][]{white2020study,white2021bananas,kadlecova2024surprisingly}.

\textbf{Eval protocol:} We evaluate a label-constrained setting with scarce ground-truth evaluations. Specifically, we use a 30-iteration predictor-guided NAS protocol, starting from $n=10$ ground-truth labeled architectures and $N=1000$ ZCP-only architectures. At each iteration, we sample 100 candidates and evaluate the top five according to the respective acquisition rule. We repeat each search 30 times and report mean performance $\pm$ standard deviation. Search quality is measured by final incumbent accuracy and the normalized area under the best-so-far accuracy curve (AUC), which captures how quickly strong architectures are identified. We additionally report mean and median AUC ranks across the evaluated benchmark--dataset--ZCP conditions. Details are in Appendix~\ref{sec:experimental_details}.

\section{Results}
\vspace{-2mm}

\begin{figure}[tbp]
    \centering
    \includegraphics[width=\linewidth]{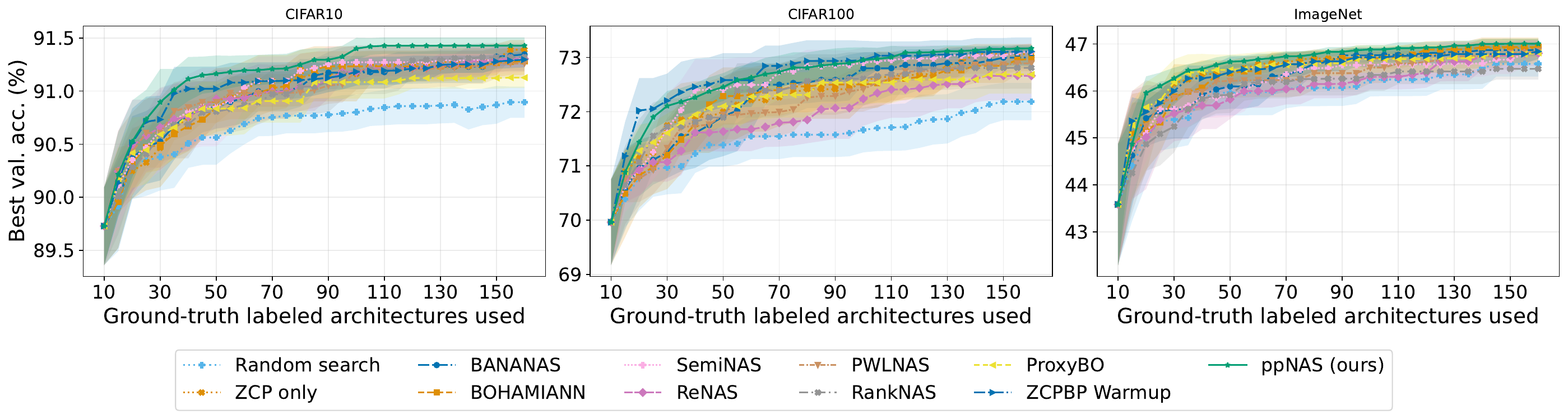}
    \vspace{-6mm}
    \caption{\textbf{Main NB201 search results} (with $n=10$ initial ground-truth-labeled architectures and the \texttt{nwot} ZCP). Curves show best-so-far validation accuracy as a function of consumed ground-truth evaluations. \framework reaches strong incumbents early in the search, and can reach a higher validation accuracy several iterations earlier.}
    \label{fig:nas_main_results}
\end{figure}

$\bullet$\,\textbf{Main results.} We first benchmark our proposed \framework against the baselines under the same limited ground-truth evaluation budget, so that comparison is \emph{fair}. Results are in Appendix~\ref{tab:nas_main_results_summary}\ref{fig:nas_main_results}, with further search trajectories in \Cref{sec:extended_main_results}. We find: \textbf{(1)} Our \framework performs best, reaching a $1.5$ median rank versus $3$ for the second-best method. \textbf{(2)} \framework consistently reaches a higher AUC than baselines, implying that its ZCP-based supervision can help find better architectures early in the search. $\Rightarrow$ \textbf{Takeaway}: \emph{Our \framework improves label-efficient NAS search}.

\begin{table}[tbp]
    \centering
     \small
     \caption{\textbf{Aggregated NB201 search results} (with $n=10$). We report mean final incumbent accuracy and normalized area under the incumbent-accuracy curve (AUC), together with AUC-based mean and median ranks across matched search runs. Best-per-column in bold. Our proposed \framework reaches the highest AUC and lowest (i.e., best) ranks.}
    \label{tab:nas_main_results_summary}
    \vspace{-5pt}
    \adjustbox{max width=\textwidth}{%
    \begin{tabular}{lllllll|ll}
\toprule
& \multicolumn{2}{c}{CIFAR10} & \multicolumn{2}{c}{CIFAR100} & \multicolumn{2}{c}{ImageNet} & \multicolumn{2}{c}{Rank $\downarrow$} \\
Method &  Final Acc & AUC & Final Acc & AUC & Final Acc. & AUC & Mean & Median  \\
\midrule
Random search & 90.89 $\pm$ 0.20 & 90.68 $\pm$ 0.28 & 72.19 $\pm$ 0.46 & 71.52 $\pm$ 0.53 & 46.58 $\pm$ 0.36 & 46.07 $\pm$ 0.72 & 9.44 & 10.00 \\
ProxyBO & 90.99 $\pm$ 0.29 & 90.80 $\pm$ 0.27 & 72.56 $\pm$ 0.56 & 72.14 $\pm$ 0.51 & 46.52 $\pm$ 0.43 & 46.15 $\pm$ 0.49 & 7.83 & 8.50 \\
ZCP only & 90.94 $\pm$ 0.40 & 90.73 $\pm$ 0.37 & 72.48 $\pm$ 0.60 & 72.03 $\pm$ 0.57 & 46.61 $\pm$ 0.35 & 46.20 $\pm$ 0.42 & 7.56 & 9.50 \\
RankNAS & 91.32 $\pm$ 0.18 & 90.96 $\pm$ 0.21 & 72.81 $\pm$ 0.51 & 72.30 $\pm$ 0.49 & 46.47 $\pm$ 0.30 & 46.02 $\pm$ 0.38 & 7.22 & 7.50 \\
ReNAS & 91.30 $\pm$ 0.15 & 90.99 $\pm$ 0.21 & 72.66 $\pm$ 0.61 & 71.93 $\pm$ 0.47 & 46.80 $\pm$ 0.37 & 46.07 $\pm$ 0.47 & 7.11 & 8.00 \\
PWLNAS & 91.26 $\pm$ 0.19 & 90.98 $\pm$ 0.19 & 72.89 $\pm$ 0.19 & 72.15 $\pm$ 0.31 & 46.85 $\pm$ 0.25 & 46.26 $\pm$ 0.37 & 5.94 & 6.00 \\
BANANAS & 91.34 $\pm$ 0.10 & 90.97 $\pm$ 0.22 & 73.00 $\pm$ 0.27 & 72.30 $\pm$ 0.37 & 46.87 $\pm$ 0.16 & 46.33 $\pm$ 0.29 & 5.28 & 5.50 \\
BOHAMIANN & 91.28 $\pm$ 0.17 & 90.98 $\pm$ 0.14 & 72.97 $\pm$ 0.37 & 72.23 $\pm$ 0.43 & \textbf{46.93 $\pm$ 0.27} & 46.42 $\pm$ 0.26 & 4.94 & 5.00 \\
ZCPBP Warmup & 91.16 $\pm$ 0.30 & 90.88 $\pm$ 0.28 & 72.70 $\pm$ 0.58 & 72.25 $\pm$ 0.57 & 46.81 $\pm$ 0.29 & 46.40 $\pm$ 0.35 & 4.94 & 4.00 \\
SemiNAS & 91.31 $\pm$ 0.16 & 91.03 $\pm$ 0.23 & \textbf{73.10 $\pm$ 0.20} & 72.52 $\pm$ 0.33 & 46.84 $\pm$ 0.29 & 46.20 $\pm$ 0.28 & 3.89 & 3.00 \\
\midrule
ppNAS (ours) & \textbf{91.36 $\pm$ 0.19} & \textbf{91.11 $\pm$ 0.18} & 73.06 $\pm$ 0.32 & \textbf{72.57 $\pm$ 0.41} & 46.87 $\pm$ 0.27 & \textbf{46.45 $\pm$ 0.34} & \textbf{1.83} & \textbf{1.50} \\
\bottomrule
\end{tabular}
} 
\vspace{-5mm}
\end{table}

\emph{What explains the improved performance?} \framework differs from baselines in several ways. Here, ReNAS and PWLNAS learn from ground-truth ranking using hinge- and logistic-ranking objectives, respectively, but do \emph{not} exploit the ZCP-only sample. Further, ZCPBP Warmup uses ZCP-induced rankings to warm up a predictor, but not afterwards. In contrast to these ranking-based approaches, \framework uses the ground-truth and ZCP-induced ranking supervision \emph{together throughout the entire fitting}. Importantly, we do not treat the proxy rankings as ground truth: their contribution is explicitly rectified using the proxy loss on the gold-labeled sample. The strong performance of \framework thus demonstrates that abundant ZCP-induced rankings and (scarce) ground-truth rankings are best combined in a principled way.

\begin{figure}[t]
    \centering
    \includegraphics[width=\linewidth,
    trim=0 10mm 0 0,
    clip]{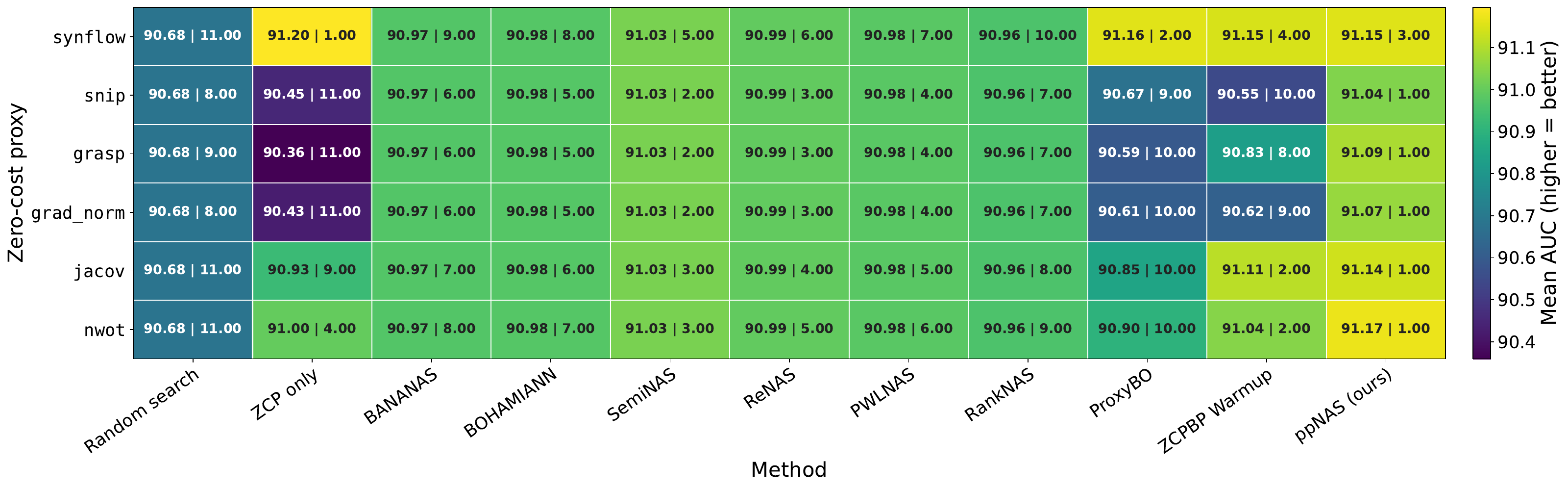}
    \vspace{-15pt}
    \caption{\textbf{Breakdown by ZCPs} (on NB201/CIFAR-10). Each cell reports mean AUC $\mid$ mean AUC rank, color is determined by the AUC. Note that ZCP-independent methods have identical values across rows. \framework ranks first except for \texttt{synflow}.}
    \vspace{-0.7cm}
    \label{fig:per_proxy_results}
\end{figure}
$\bullet$\,\textbf{Breakdown by ZCPs.} We report mean AUC together with mean AUC rank for every ZCP and method for CIFAR10 in \Cref{fig:per_proxy_results}; further results are in Appendix~\ref{sec:per_proxy_results_extended}. Across all available NB201 ZCP--dataset combinations, \framework consistently ranks first or within the three best methods, except for \texttt{synflow} on ImageNet (4th rank). $\Rightarrow$ \textbf{Takeaway}: \emph{ \framework works with all ZCP-induced supervisions.}

$\bullet$\,\textbf{Sensitivity: different starting budgets.} We repeat our main experiment but evaluate $n\in\{5,20,50\}$ initial samples. Results are in Appendix~\ref{sec:different_initial_sizes}. We observe: \textbf{(1)}~\framework maintains the best aggregate rank across all settings (with shared first rank with BANANAS for $n=50$). \textbf{(2)}~Differences to the other baselines reduce as more initial samples are available. This is expected: with more initial samples, stronger predictors can be learned from the ground-truth supervision alone.

$\bullet$\,\textbf{Sensitivity: different unlabeled sizes.} We now vary the size of $\UnlabeledDataset$ across $N \in \{50, 100, 250, 500, 1000\,\text{(default)},2000,3000\}$ for \framework. Results are in  Appendix~\ref{sec:unlabeled_ablation}. We observe that performance is relatively stable across the budgets. $\Rightarrow$ \textbf{Takeaway}: \emph{\framework can exploit $\UnlabeledDataset$ across different sample sizes.}

\begin{wrapfigure}{r}{0.5\textwidth}
  \centering
  \vspace{-0.2cm}
    \includegraphics[width=\linewidth,trim=0 2mm 0 0,
    clip]{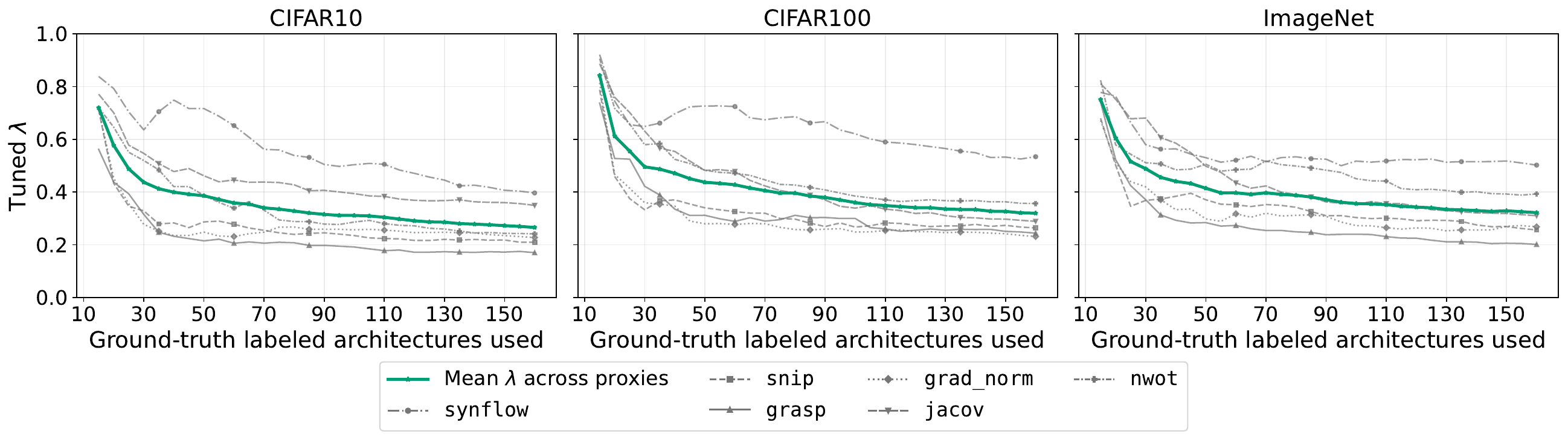}
  \caption{Change of $\lambda$ over NAS iterations.}\label{fig:lambda_analysis}
  \vspace{-10pt}
\end{wrapfigure}

$\bullet$\,\textbf{Analysis of hyperparameter $\lambda$.}
In \Cref{fig:lambda_analysis}, we visualize how $\lambda$ evolves over the search process; further results in Appendix\ref{sec:lambda_analysis}, \Cref{fig:lambda_analysis_over_time}. We consistently observe that, as more ground-truth evaluations become available, $\lambda$ decreases from an initial high $\sim$0.8 and stabilizes around $0.2$--$0.3$. Further, the initial $\lambda$ is higher when fewer ground-truth-labeled architectures are available. This is consistent with the intended role of $\lambda$: when ground-truth supervision is scarce, the model places greater weight on the additional ZCP-induced ranking supervision. $\Rightarrow$ \textbf{Takeaway}: \emph{\framework is effective in adapting $\lambda$ over the search.}

$\bullet$\,\textbf{Robustness: different architecture encoding.} We repeat our main experiment and use \texttt{arch2vec} representations \citep{yan2020does} instead. The results are in Appendix~\ref{sec:other:encoding}. \framework again reaches the best rank. Further, the first four ranks are all held by ranking-based methods, suggesting that \texttt{arch2vec} constructs information beneficial for (supervised) ranking. $\Rightarrow$ \textbf{Takeaway}: \emph{Our results are robust to different architecture encoding.}

$\bullet$\,\textbf{Robustness: different search spaces.} We repeat the main experiment on the NB101 and TNB benchmarks. Results are reported in \Cref{sec:different_search_spaces}. We find that \framework again achieves the best mean and median ranks on both benchmarks. The findings for TNB mean that \framework also improves performance for macro-level search spaces. $\Rightarrow$ \textbf{Takeaway}: \emph{Our results are robust to different search spaces.}

$\bullet$\,\textbf{Stress test: misspecified ZCP direction.}
Throughout our experiments, we use the common convention that larger ZCP scores indicate better architectures (see \citet{abdelfattah2021zero,Krishnakumar2022zerocost} and \Cref{fig:nb201_proxy_correlation}). In practice, a ZCP with the opposite direction could simply be reverse-coded. Here, however, we deliberately leave the direction unchanged to stress-test whether \framework can automatically remain robust to such misspecified ZCP supervision. We now use NB101 to stress-test what happens when ZCP direction is misspecified. Here, for \texttt{snip}, \texttt{grad\_norm}, and \texttt{jacov}, larger ZCP scores are associated with lower validation performance, as indicated by their negative Spearman correlations in \Cref{fig:nb101_proxy_correlation}. 

Despite the ZCP misspecification, \framework remains competitive. In the per-ZCP comparison (\Cref{fig:nb101_heatmap}), \framework ranks first with \texttt{snip} and \texttt{jacov}, second with \texttt{grad\_norm}, and first for all remaining ZCPs. This stress test illustrates that the learned architecture ranking can remain robust even when the direction of the auxiliary ZCP supervision is incorrectly specified. $\Rightarrow$ \textbf{Takeaway}: \emph{\framework is robust to misspecified ZCP supervision. More broadly, this reflects the role of the prediction-powered formulation, where ZCP-induced preferences provide additional supervision but are not assumed to be perfectly reliable.}

$\bullet$\,\textbf{Runtimes.}
Runtimes for \framework are comparable to the baselines; see \Cref{fig:runtimes}.

\textbf{Conclusion.} Our \framework is the first prediction-powered approach for label-efficient NAS, by combining scarce ground-truth supervision with large-scale ZCP-induced rankings through a rectified ranking objective. Across our experiments, it achieves consistent performance improvements across datasets, search spaces, and ZCPs. Future work could extend our prediction-powered approach to LLMs and study whether similar ideas can support the study of scaling laws.

\bibliography{bib.bib}
\bibliographystyle{bib_style}

\appendix

\newpage
\section{Background on PPI}\label{sec:ppi_recap}

We provide a brief overview of PPI \citep{Angelopoulos2023ppi,angelopoulos2024ppiefficientpredictionpoweredinference}. Let $(X,Y)\sim P_{XY}$, where $X\in\mathcal{X}$ denotes the input and $Y\in\mathcal{Y}$ the corresponding ground-truth label. We observe: (a)~a small labeled sample $S_n= \{( X_1,Y_1),\ldots,(X_n,Y_n) \}$, with $(X_i,Y_i) \overset{\mathrm{iid}}{\sim} \mathbb{P}_{XY}$; and (b)~a larger, unlabeled sample $\widetilde{S}_N=\{\widetilde{X}_1,\ldots,\widetilde{X}_N\}$, with $n \ll N$ and $\widetilde{X}_i\overset{\mathrm{iid}}{\sim}P_X$, where $P_X$ denotes the marginal distribution of $X$ under $P_{XY}$. The corresponding ground-truth labels are unobserved for (b). We additionally assume access to a pre-trained predictor $f:\mathcal{X}\rightarrow\mathcal{Y}$ to generate noisy, auxiliary labels. 

\textbf{PPI objective.}
For a loss function $\ell_\theta$, the goal is to learn a target parameter
\begin{equation}
    \theta^*
    =
    \arg\min_{\theta\in\Theta}
    \mathbb{E}\left[
        \ell_\theta(X,Y)
    \right].
\end{equation}
To do so, we define the empirical ground-truth loss $L_n(\theta)$ and the prediction-based losses $\widetilde{L}_N^f(\theta)$: 
\begin{align}\label{eq:ppi_loss_terms}
    L_n(\theta) = \frac{1}{n}\sum_{i=1}^{n}\ell_\theta(X_i,Y_i),
    \quad
    L_n^f(\theta) = \frac{1}{n}\sum_{i=1}^{n}\ell_\theta\!\left(X_i,f(X_i)\right), \quad
    \widetilde{L}_N^f(\theta) = \frac{1}{N}\sum_{i=1}^{N}\ell_\theta\!\left(\widetilde{X}_i,f(\widetilde{X}_i)\right).
\end{align}
PPI estimates $\theta^*$ through the \textit{rectified} prediction-based objective
\begin{equation}
\label{eq:ppi_objective}
    \widehat{\theta}^{\mathrm{PPI}}
    =
    \arg\min_{\theta}
    \bigg\{
        L_n(\theta)
        +
        \underbrace{\big[
            \widetilde{L}_N^f(\theta)
            -
            L_n^f(\theta)
        \big]}_{\text{rectifier}}
    \bigg\} .
\end{equation}
The first term exploits the auxiliary predictions from the large unlabeled sample, while the so-called \textit{rectifier} compares prediction-based and ground-truth losses on the same labeled observations and thereby corrects systematic errors introduced by a noisy predictor $f$.

\newpage
\section{Experimental details}\label{sec:experimental_details}
\textbf{NAS search protocol.} All methods are evaluated within the same iterative predictor-guided NAS protocol. Each search starts from $n=10$ randomly selected architectures with observed ground-truth validation performance. We additionally provide $N=1000$ architectures with ZCP information but without ground-truth performance labels. The unlabeled sample size is kept fixed throughout the search. 

For each search iteration $t=1,\ldots,30$, the corresponding performance predictor is fitted using the information available at that iteration. We then sample 100 candidate architectures from the remaining search space, score them according to the method-specific acquisition rule, and select the top five for ground-truth evaluation. The newly evaluated architectures are added to the labeled set before the next predictor fit. Thus, in the main setting, the labeled set grows from $10$ to $160$ architectures over the 30 search iterations. For methods that require ensemble-based uncertainty, we use an ensemble of three independently fitted predictors. PPNAS instead uses the parameter covariance estimated by its ranking model for uncertainty-aware acquisition and therefore requires only a single fitted model.

We repeat every search configuration 30 times. At each iteration, we record the validation accuracy of the best architecture observed so far. We summarize each search trajectory using two complementary metrics: final incumbent accuracy at the end of the search and the normalized area under the best-so-far accuracy curve (AUC). We compute AUC over the common 30-iteration search horizon using trapezoidal integration and normalize by the horizon length, such that AUC remains on the same numerical scale as validation accuracy. Final accuracy and AUC are reported as the mean and standard deviation across repeated searches.

To summarize performance across experimental conditions, we rank methods by their AUC within each benchmark--dataset--proxy condition and report the mean and median rank across these comparisons. AUC rewards methods that identify strong architectures early in the search, whereas final accuracy reflects the quality of the incumbent after the complete evaluation budget has been consumed.

\textbf{Baselines.} ProxyBO and ZCPBP Warmup are implementations based on the respective papers, as we could not find publicly usable implementations.

\newpage
\section{Extended main results}\label{sec:extended_main_results}
\begin{figure}[htbp]
    \centering
    \includegraphics[width=\linewidth]{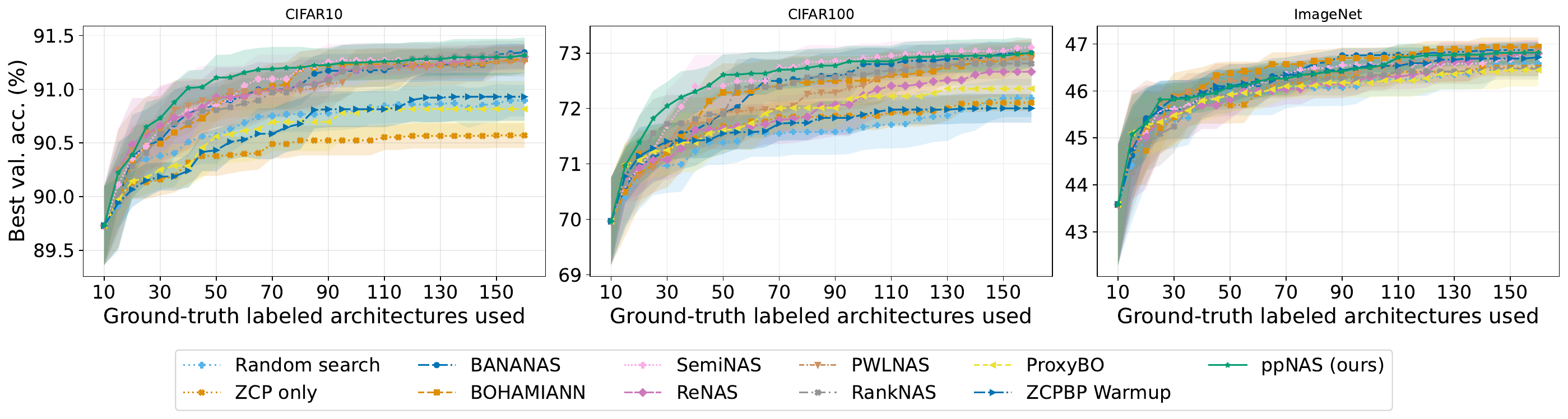}
    \caption{Search trajectory on NB201 using \texttt{grad\_norm}.}
    \label{fig:nb201_grad_norm}
\end{figure}

\begin{figure}[htbp]
    \centering
    \includegraphics[width=\linewidth]{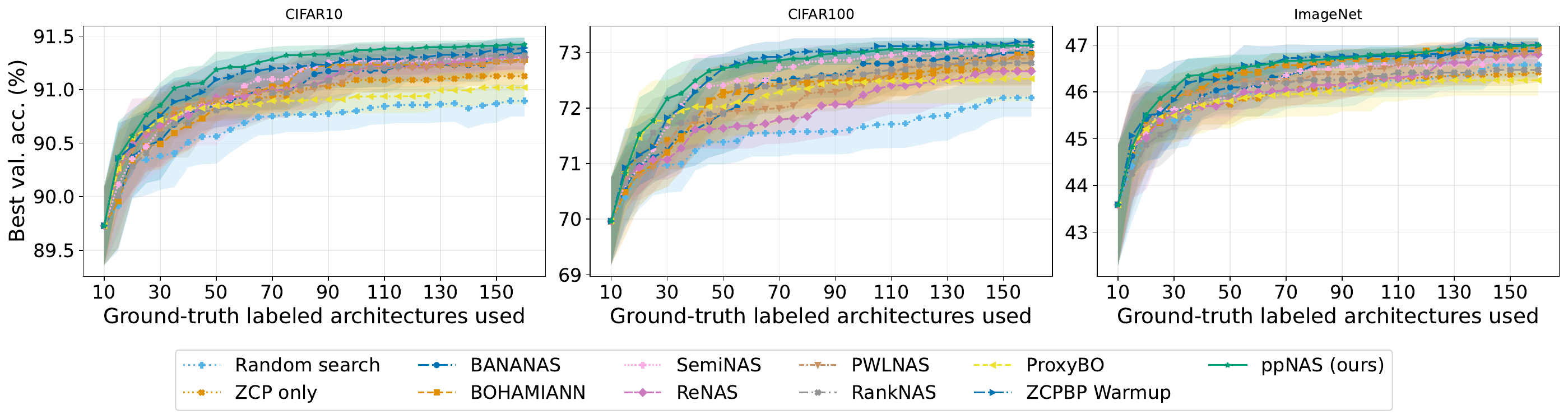}
    \caption{Search trajectory on NB201 using \texttt{jacov} (Jacobian covariance).}
    \label{fig:nb201_jacov}
\end{figure}

\begin{figure}[htbp]
    \centering
    \includegraphics[width=\linewidth]{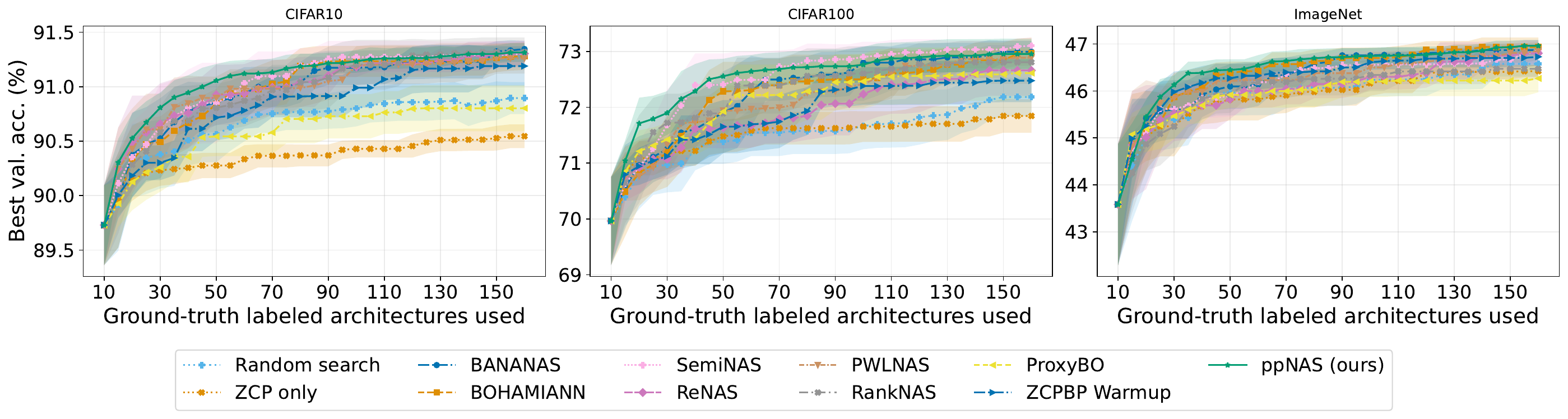}
    \caption{Search trajectory on NB201 using \texttt{grasp}.}
    \label{fig:nb201_nwot}
\end{figure}

\begin{figure}[htbp]
    \centering
    \includegraphics[width=\linewidth]{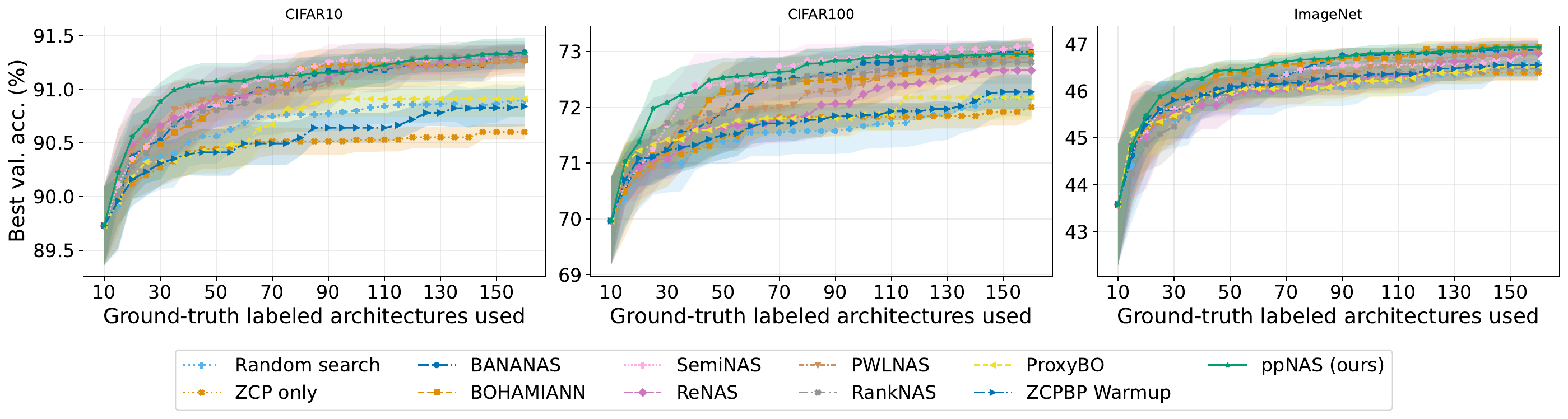}
        \caption{Search trajectory on NB201 using \texttt{snip}.}
    \label{fig:nb201_snip}
\end{figure}

\begin{figure}[htbp]
    \centering
    \includegraphics[width=\linewidth]{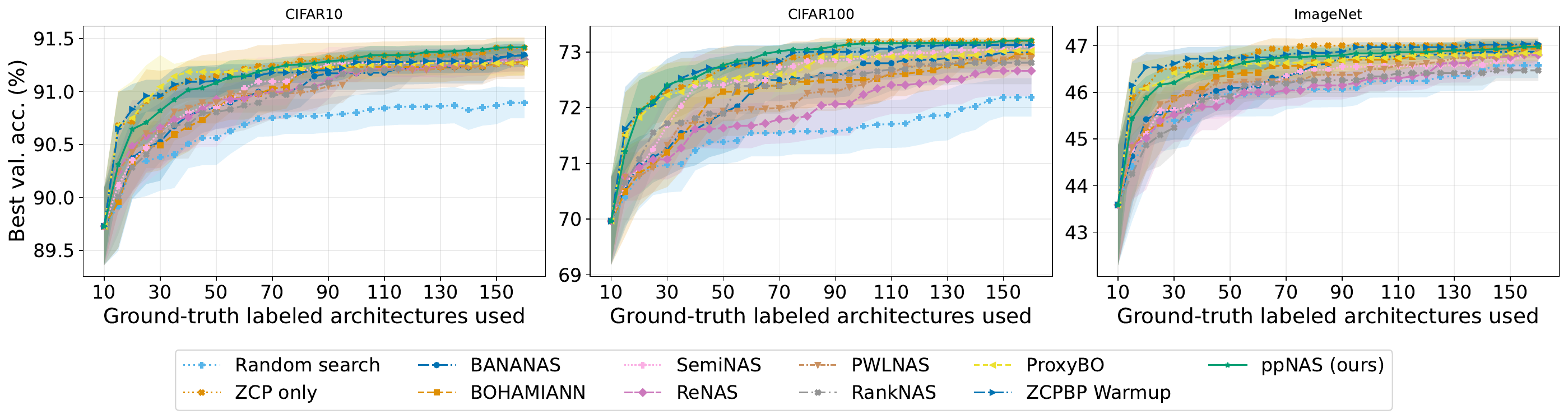}
    \caption{Search trajectory on NB201 using \texttt{synflow}.}
    \label{fig:nb201_synflow}
\end{figure}

We here visualize the search trajectories for other ZCPs. Visualizations are in \Cref{fig:nb201_grad_norm,fig:nb201_jacov,fig:nb201_nwot,fig:nb201_snip,fig:nb201_synflow}.

\clearpage
\section{$\mathbf{\lambda}$ analysis}\label{sec:lambda_analysis}
\begin{figure}[htbp]
    \centering
    \includegraphics[width=0.8\linewidth]{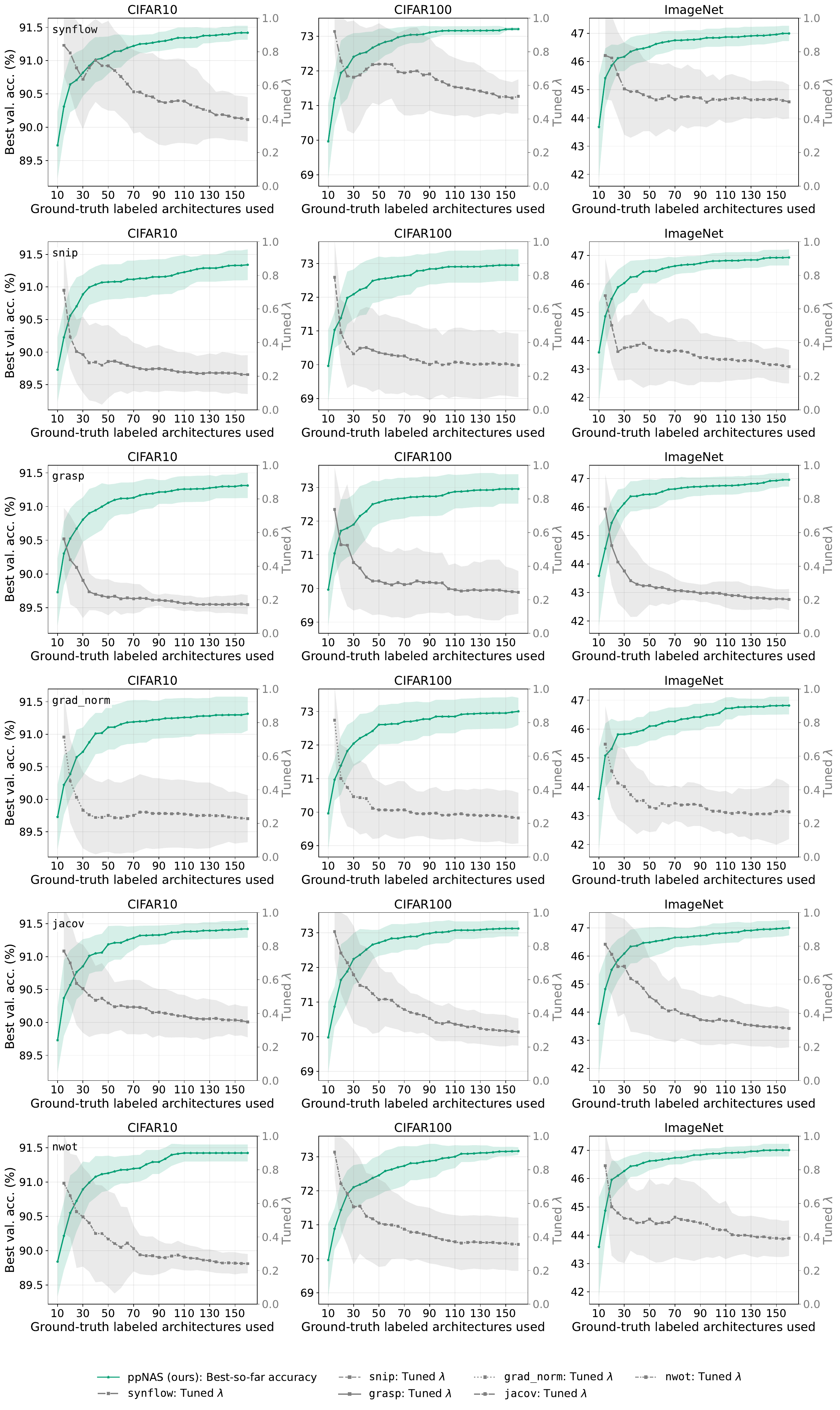}
    \caption{$\lambda$ as a function of the number of ground-truth-labeled architecture acquired during NAS iterations.}
    \label{fig:extended_lambda_analysis}
\end{figure}

\begin{figure}[htp]
\centering
\begin{subfigure}{0.8\textwidth}
  \centering
  \includegraphics[width=\linewidth]{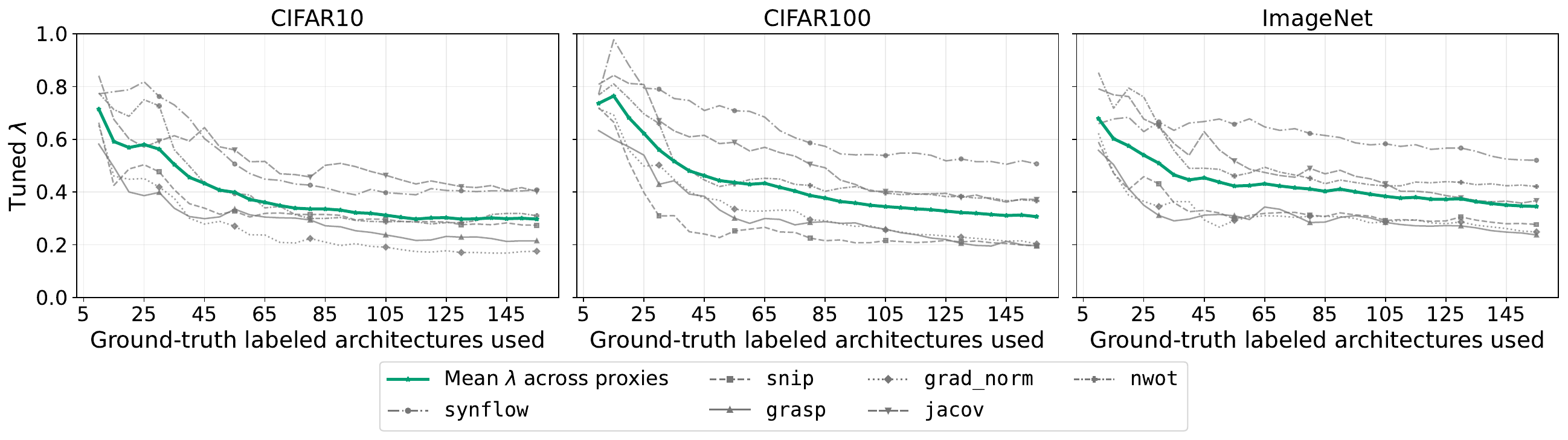}
  \caption{5 starting points}
  \label{subfig:fig1}
\end{subfigure}

\begin{subfigure}{0.8\textwidth}
  \centering
  \includegraphics[width=\linewidth]{plots/rankppi_lambda_evolution_nb201_rappi_hajek_proxy_average_gold10_gold_labels_cap30.pdf}
  \caption{10 starting points}
  \label{subfig:fig1}
\end{subfigure}

\begin{subfigure}{0.8\textwidth}
  \centering
  \includegraphics[width=\linewidth]{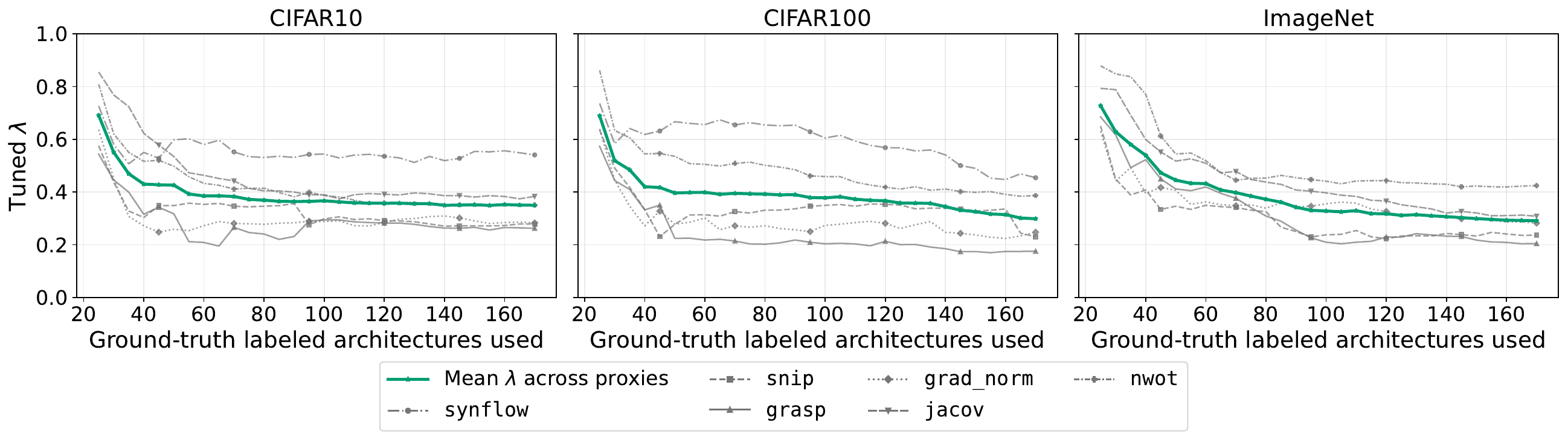}
  \caption{20 starting points}
  \label{subfig:fig2}
\end{subfigure}

\begin{subfigure}{0.8\textwidth}
  \centering
  \includegraphics[width=\linewidth]{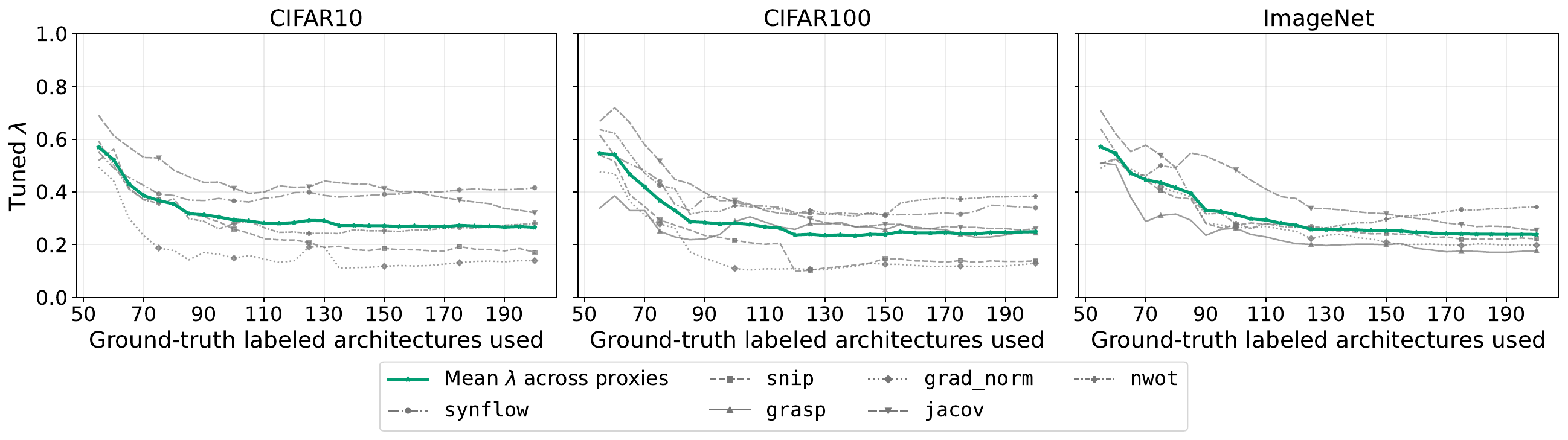}
  \caption{50 starting points}
  \label{subfig:fig2}
\end{subfigure}

\caption{Evolution of $\lambda$ over time.}
\label{fig:lambda_analysis_over_time}
\end{figure}

We here provide extended visualizations for the behaviour of the $\lambda$ weighting parameter over time, see \Cref{fig:extended_lambda_analysis}. We observe a clear pattern, where the influence of the ZCP rectification is highest in the early iterations, as given by a high $\lambda$. As more ground-truth labels become available, the weight slowly decays until it settles around $0.3$. This behavior is consistent for other starting conditions; see \Cref{fig:lambda_analysis_over_time}.

\pagebreak
\newpage
\section{Different initial sizes}\label{sec:different_initial_sizes}
\begin{figure}[!htbp]
    \centering
    \includegraphics[width=\linewidth]{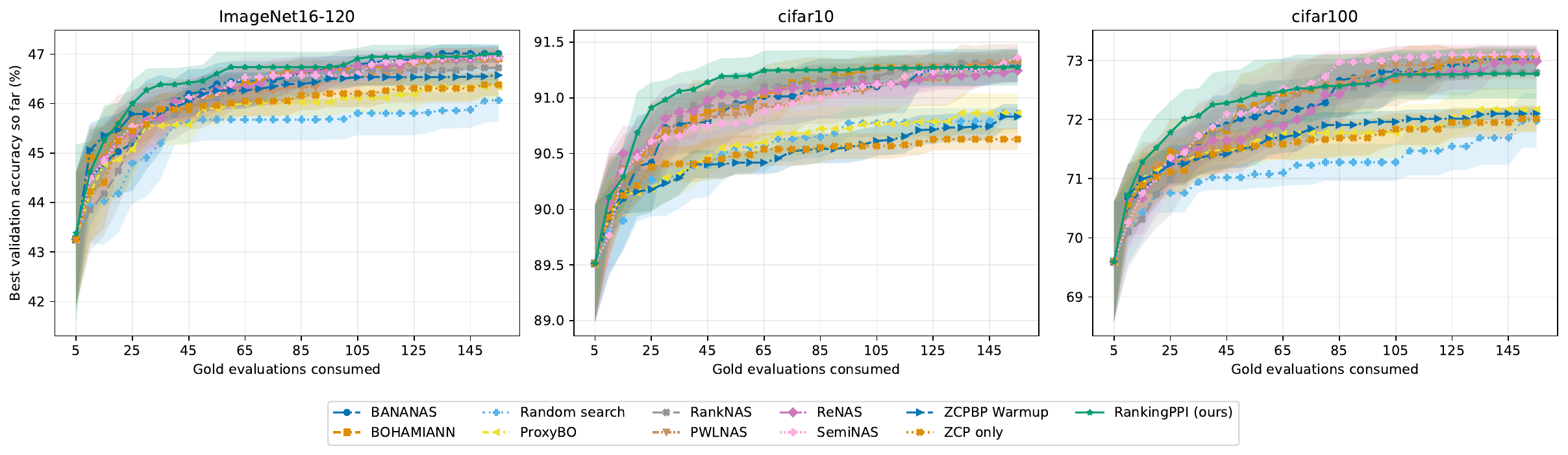}
    \caption{grad norm, gold = 5}
    \label{fig:gold_5_example}
\end{figure}

In the main experiments, we use $n=10$ initial samples to train the performance predictors prior to the first NAS iterations. We here repeat this experiment, again on NB201, but use $n\in\{5,20,50\}$. The results are in \Cref{tab:nb201_n_5,tab:nb201_n_20,tab:nb201_n_50}, with an exemplary search trajectory in \Cref{fig:gold_5_example} for the case of $n=5$. We observe that at each level of initially labeled samples, our \framework outperforms the baselines and reaches the highest final accuracy, AUC, and best ranks.

\begin{table}[ht]
    \centering
    \small
    \caption{Ranking across all search experiments on NB201, for $5$ initial samples.}
    \label{tab:nb201_n_5}
    \adjustbox{max width=\linewidth}{
    \begin{tabular}{lllllll|ll}
\toprule
& \multicolumn{2}{c}{CIFAR10} & \multicolumn{2}{c}{CIFAR100} & \multicolumn{2}{c}{ImageNet} & \multicolumn{2}{c}{Rank $\downarrow$} \\
Method &  Final Acc & AUC & Final Acc & AUC & Final Acc. & AUC & Mean & Median  \\
\midrule
Random search & 90.86 $\pm$ 0.08 & 90.55 $\pm$ 0.19 & 71.98 $\pm$ 0.61 & 71.18 $\pm$ 0.28 & 46.07 $\pm$ 0.57 & 45.46 $\pm$ 0.51 & 10.72 & 11.00 \\
ProxyBO & 91.09 $\pm$ 0.26 & 90.81 $\pm$ 0.26 & 72.52 $\pm$ 0.62 & 72.14 $\pm$ 0.64 & 46.54 $\pm$ 0.39 & 46.12 $\pm$ 0.42 & 7.11 & 8.50 \\
PWLNAS & 91.31 $\pm$ 0.23 & 90.91 $\pm$ 0.23 & 72.97 $\pm$ 0.38 & 72.23 $\pm$ 0.23 & 46.89 $\pm$ 0.25 & 46.23 $\pm$ 0.31 & 7.06 & 7.00 \\
ZCP only & 90.93 $\pm$ 0.41 & 90.73 $\pm$ 0.35 & 72.52 $\pm$ 0.63 & 72.09 $\pm$ 0.62 & 46.64 $\pm$ 0.35 & 46.18 $\pm$ 0.42 & 7.00 & 8.50 \\
ReNAS & 91.25 $\pm$ 0.14 & 90.98 $\pm$ 0.17 & 72.99 $\pm$ 0.29 & 72.12 $\pm$ 0.32 & 46.95 $\pm$ 0.28 & 46.28 $\pm$ 0.28 & 6.56 & 6.50 \\
RankNAS & 91.31 $\pm$ 0.10 & 90.99 $\pm$ 0.10 & 72.80 $\pm$ 0.38 & 72.15 $\pm$ 0.42 & 46.72 $\pm$ 0.33 & 46.09 $\pm$ 0.37 & 6.39 & 7.00 \\
SemiNAS & \textbf{91.36 $\pm$ 0.16} & 90.90 $\pm$ 0.19 & \textbf{73.11 $\pm$ 0.21} & 72.38 $\pm$ 0.30 & 46.94 $\pm$ 0.23 & 46.28 $\pm$ 0.40 & 5.44 & 5.00 \\
BANANAS & 91.29 $\pm$ 0.20 & 90.92 $\pm$ 0.29 & 73.02 $\pm$ 0.29 & 72.23 $\pm$ 0.40 & \textbf{47.01 $\pm$ 0.26} & 46.31 $\pm$ 0.41 & 4.94 & 5.00 \\
ZCPBP Warmup & 91.12 $\pm$ 0.34 & 90.87 $\pm$ 0.35 & 72.70 $\pm$ 0.56 & 72.24 $\pm$ 0.60 & 46.82 $\pm$ 0.29 & 46.44 $\pm$ 0.35 & 4.78 & 4.00 \\
BOHAMIANN & 91.33 $\pm$ 0.12 & 90.98 $\pm$ 0.23 & 73.07 $\pm$ 0.19 & 72.30 $\pm$ 0.46 & 46.91 $\pm$ 0.29 & 46.29 $\pm$ 0.29 & 4.11 & 3.50 \\
\midrule
ppNAS (ours) & 91.33 $\pm$ 0.20 & \textbf{91.10 $\pm$ 0.21} & 73.00 $\pm$ 0.39 & \textbf{72.45 $\pm$ 0.38} & 46.93 $\pm$ 0.26 & \textbf{46.46 $\pm$ 0.28} & \textbf{1.89} & \textbf{2.00} \\
\bottomrule
\end{tabular}
}
\end{table}

\begin{table}[ht]
    \centering
    \caption{Ranking across all search experiments on NB201, for $20$ initial samples.}
    \label{tab:nb201_n_20}
    \adjustbox{max width=\linewidth}{
   \begin{tabular}{lllllll|ll}
\toprule
& \multicolumn{2}{c}{CIFAR10} & \multicolumn{2}{c}{CIFAR100} & \multicolumn{2}{c}{ImageNet} & \multicolumn{2}{c}{Rank $\downarrow$} \\
Method &  Final Acc & AUC & Final Acc & AUC & Final Acc. & AUC & Mean & Median  \\
\midrule
Random search & 90.95 $\pm$ 0.16 & 90.78 $\pm$ 0.18 & 72.01 $\pm$ 0.56 & 71.40 $\pm$ 0.34 & 46.03 $\pm$ 0.46 & 45.71 $\pm$ 0.54 & 10.56 & 11.00 \\
ZCP only & 90.95 $\pm$ 0.38 & 90.78 $\pm$ 0.32 & 72.44 $\pm$ 0.63 & 72.01 $\pm$ 0.61 & 46.66 $\pm$ 0.40 & 46.28 $\pm$ 0.39 & 7.75 & 9.50 \\
ProxyBO & 91.07 $\pm$ 0.28 & 90.85 $\pm$ 0.28 & 72.33 $\pm$ 0.66 & 72.05 $\pm$ 0.60 & 46.58 $\pm$ 0.38 & 46.24 $\pm$ 0.44 & 7.69 & 9.00 \\
RankNAS & 91.27 $\pm$ 0.17 & 91.00 $\pm$ 0.22 & 72.76 $\pm$ 0.60 & 72.12 $\pm$ 0.53 & 46.76 $\pm$ 0.25 & 46.18 $\pm$ 0.15 & 7.25 & 8.00 \\
ReNAS & \textbf{91.38 $\pm$ 0.15} & 91.00 $\pm$ 0.12 & 72.81 $\pm$ 0.42 & 72.09 $\pm$ 0.37 & 46.66 $\pm$ 0.43 & 46.30 $\pm$ 0.39 & 6.69 & 6.00 \\
ZCPBP Warmup & 91.18 $\pm$ 0.32 & 90.90 $\pm$ 0.30 & 72.77 $\pm$ 0.55 & 72.27 $\pm$ 0.58 & 46.80 $\pm$ 0.30 & 46.42 $\pm$ 0.32 & 5.75 & 7.00 \\
BOHAMIANN & 91.32 $\pm$ 0.12 & 90.96 $\pm$ 0.15 & 72.68 $\pm$ 0.47 & 72.13 $\pm$ 0.45 & 46.92 $\pm$ 0.29 & 46.48 $\pm$ 0.24 & 5.50 & 6.00 \\
PWLNAS & 91.33 $\pm$ 0.08 & 91.08 $\pm$ 0.02 & 72.84 $\pm$ 0.35 & 71.98 $\pm$ 0.30 & 46.85 $\pm$ 0.24 & 46.35 $\pm$ 0.30 & 5.25 & 5.00 \\
BANANAS & 91.31 $\pm$ 0.23 & 91.03 $\pm$ 0.21 & 73.07 $\pm$ 0.18 & 72.37 $\pm$ 0.26 & 46.75 $\pm$ 0.20 & 46.37 $\pm$ 0.17 & 4.56 & 4.00 \\
SemiNAS & 91.33 $\pm$ 0.16 & 91.06 $\pm$ 0.10 & \textbf{73.14 $\pm$ 0.09} & 72.40 $\pm$ 0.29 & \textbf{46.93 $\pm$ 0.33} & 46.43 $\pm$ 0.34 & 3.56 & 3.00 \\
\midrule
ppNAS (ours) & 91.36 $\pm$ 0.16 & \textbf{91.15 $\pm$ 0.17} & 73.10 $\pm$ 0.25 & \textbf{72.64 $\pm$ 0.42} & 46.91 $\pm$ 0.22 & \textbf{46.54 $\pm$ 0.23} & \textbf{1.44} & \textbf{1.00} \\
\bottomrule
\end{tabular}
}
\end{table}

\begin{table}[ht]
    \centering
    \caption{Ranking across all search experiments on NB201, for $50$ initial samples.}
    \label{tab:nb201_n_50}
    \adjustbox{max width=\linewidth}{
    \begin{tabular}{lllllll|ll}
\toprule
& \multicolumn{2}{c}{CIFAR10} & \multicolumn{2}{c}{CIFAR100} & \multicolumn{2}{c}{ImageNet} & \multicolumn{2}{c}{Rank $\downarrow$} \\
Method &  Final Acc & AUC & Final Acc & AUC & Final Acc. & AUC & Mean & Median  \\
\midrule
Random search & 91.10 $\pm$ 0.20 & 90.86 $\pm$ 0.16 & 72.53 $\pm$ 0.51 & 71.99 $\pm$ 0.42 & 46.40 $\pm$ 0.43 & 46.06 $\pm$ 0.32 & 9.94 & 10.00 \\
ZCP only & 90.94 $\pm$ 0.36 & 90.83 $\pm$ 0.30 & 72.39 $\pm$ 0.63 & 72.08 $\pm$ 0.57 & 46.56 $\pm$ 0.39 & 46.30 $\pm$ 0.51 & 9.00 & 10.00 \\
ProxyBO & 91.01 $\pm$ 0.27 & 90.87 $\pm$ 0.25 & 72.44 $\pm$ 0.68 & 72.16 $\pm$ 0.58 & 46.48 $\pm$ 0.43 & 46.26 $\pm$ 0.44 & 9.00 & 10.00 \\
RankNAS & 91.24 $\pm$ 0.15 & 91.07 $\pm$ 0.18 & 72.89 $\pm$ 0.37 & 72.45 $\pm$ 0.52 & 46.67 $\pm$ 0.22 & 46.33 $\pm$ 0.29 & 7.50 & 8.00 \\
ReNAS & 91.26 $\pm$ 0.20 & 91.00 $\pm$ 0.16 & 73.01 $\pm$ 0.24 & 72.50 $\pm$ 0.28 & 46.82 $\pm$ 0.41 & 46.58 $\pm$ 0.41 & 7.25 & 7.00 \\
ZCPBP Warmup & 91.22 $\pm$ 0.27 & 90.99 $\pm$ 0.25 & 72.98 $\pm$ 0.39 & 72.48 $\pm$ 0.40 & 46.88 $\pm$ 0.23 & 46.55 $\pm$ 0.31 & 5.38 & 6.00 \\
PWLNAS & 91.33 $\pm$ 0.17 & 91.15 $\pm$ 0.12 & 72.95 $\pm$ 0.33 & 72.51 $\pm$ 0.43 & 46.89 $\pm$ 0.36 & 46.59 $\pm$ 0.36 & 4.94 & 5.00 \\
SemiNAS & 91.31 $\pm$ 0.19 & 91.06 $\pm$ 0.23 & 72.91 $\pm$ 0.54 & 72.59 $\pm$ 0.56 & \textbf{47.01 $\pm$ 0.24} & 46.65 $\pm$ 0.28 & 4.75 & 4.50 \\
BOHAMIANN & 91.34 $\pm$ 0.13 & 91.13 $\pm$ 0.19 & \textbf{73.13 $\pm$ 0.22} & 72.68 $\pm$ 0.26 & 46.96 $\pm$ 0.36 & 46.65 $\pm$ 0.45 & 3.56 & 3.00 \\
BANANAS & \textbf{91.37 $\pm$ 0.07} & \textbf{91.22 $\pm$ 0.13} & 73.03 $\pm$ 0.28 & 72.61 $\pm$ 0.23 & 46.99 $\pm$ 0.25 & \textbf{46.68 $\pm$ 0.33} & \textbf{2.25} & \textbf{2.00} \\
\midrule
ppNAS (ours) & 91.36 $\pm$ 0.12 & 91.18 $\pm$ 0.14 & 73.11 $\pm$ 0.25 & \textbf{72.76 $\pm$ 0.31} & 46.92 $\pm$ 0.26 & 46.64 $\pm$ 0.33 & 2.44 & \textbf{2.00} \\
\bottomrule
\end{tabular}
}
\end{table}

\clearpage
\section{Different unlabeled sizes}\label{sec:unlabeled_ablation}
\begin{figure}[h]
    \centering
    \includegraphics[width=0.8\linewidth]{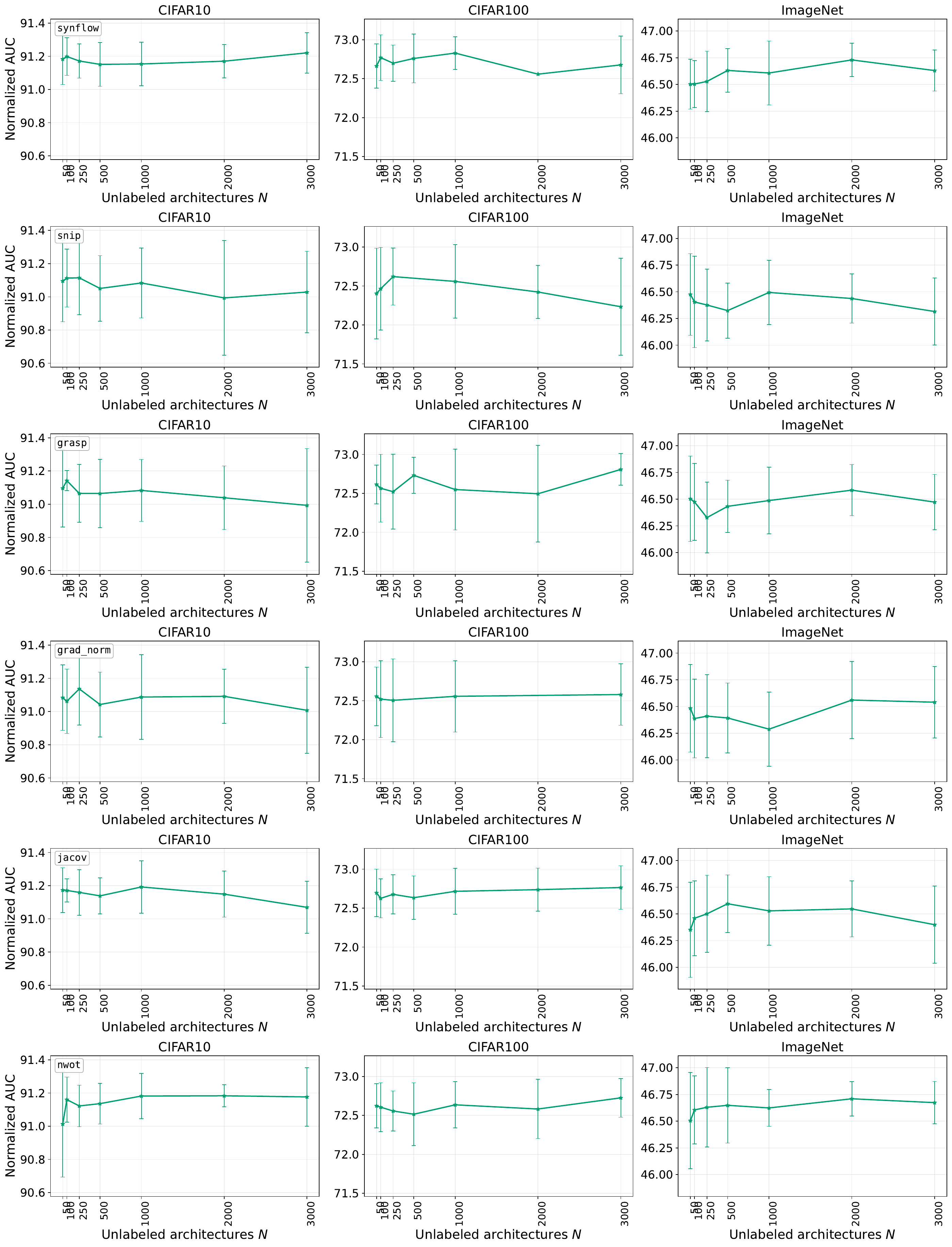}
    \caption{The effect of different $N$ on the final mean AUC.}
    \label{fig:different_unlabeled_sizes_auc}
\end{figure}

\begin{figure}[h]
    \centering
    \includegraphics[width=0.8\linewidth]{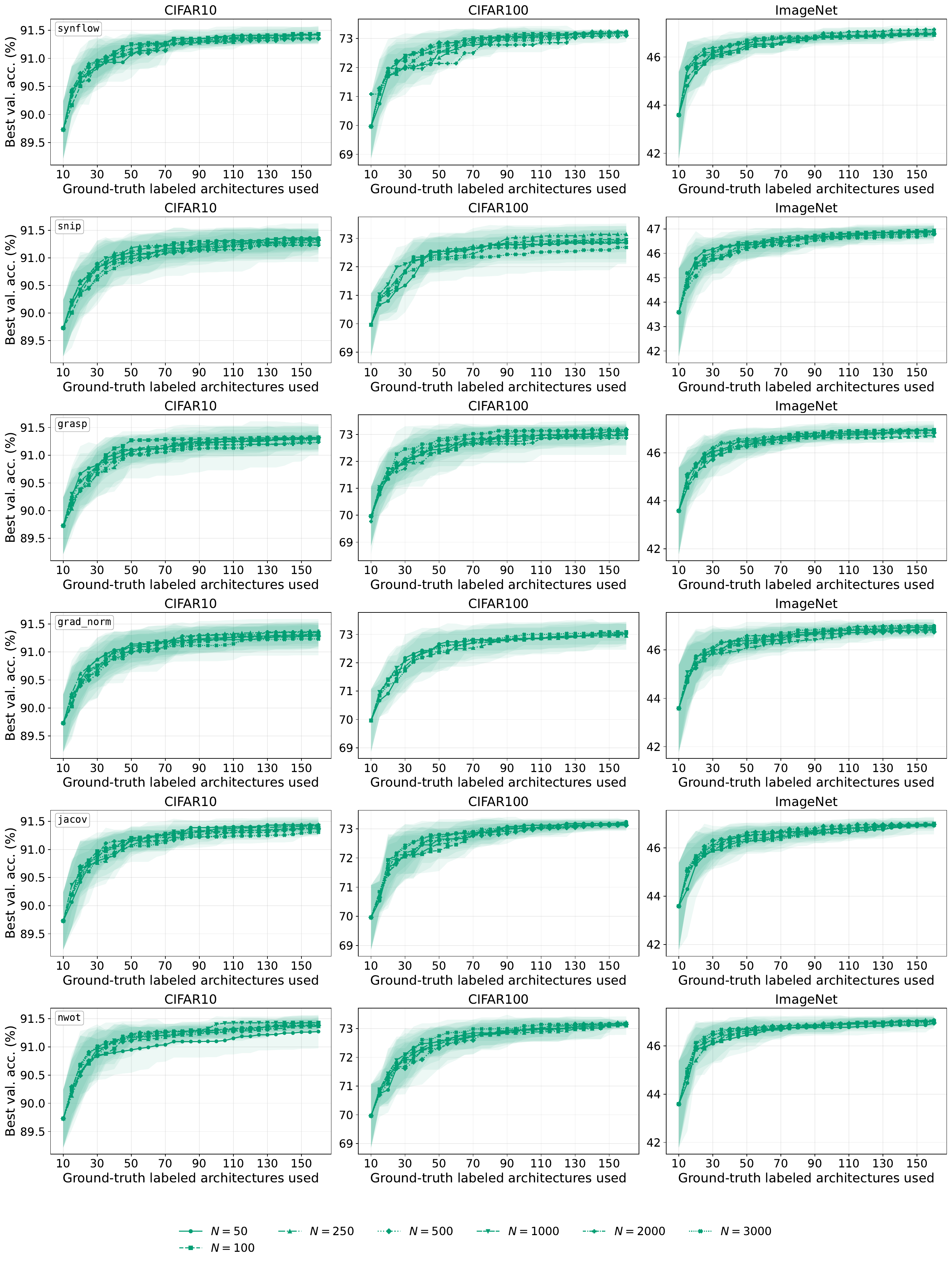}
    \caption{The effect of different $N$ on incumbent curves.}
    \label{fig:different_unlabeled_sizes_curves}
\end{figure}

In our main experiments, we use $|\UnlabeledDataset|=1000$. We now vary this for \framework to gauge its sensitivity. Results are in \Cref{fig:different_unlabeled_sizes_auc,fig:different_unlabeled_sizes_curves}. We observe that \textbf{(1)} \framework can exploit variably-sized $\UnlabeledDataset$, \textbf{(2)} search curves show similar patterns across the budgets,i.e., there is no outlier, and \textbf{(3)} final performance is only minimally affected.

\clearpage
\section{Other architecture encoding}\label{sec:other:encoding}
In the main experiments, we use a straightforward one-hot representation. We now repeat this, but use arch2vec \citep{yan2020does} to encode the neural architecture. Note that SemiNAS and ProxyBO are not compatible with this encoding type. Results are in \Cref{tab:other_encoding}. \framework again performs strongly, reaching the first place by rank.

\begin{table}[htbp]
    \centering
        \caption{Ranking across all search experiments on NB201, \emph{using arch2vec neural architecture representations} \citep{yan2020does}. Our \framework reaches the best rank.}
        \label{tab:other_encoding}
         \adjustbox{max width=\linewidth}{
    \begin{tabular}{lllllll|ll}
\toprule
& \multicolumn{2}{c}{CIFAR10} & \multicolumn{2}{c}{CIFAR100} & \multicolumn{2}{c}{ImageNet} & \multicolumn{2}{c}{Rank $\downarrow$} \\
Method &  Final Acc & AUC & Final Acc & AUC & Final Acc. & AUC & Mean & Median  \\
\midrule
Random search & 90.90 $\pm$ 0.20 & 90.68 $\pm$ 0.28 & 72.19 $\pm$ 0.46 & 71.52 $\pm$ 0.53 & 46.58 $\pm$ 0.36 & 46.07 $\pm$ 0.72 & 8.61 & 9.00 \\
ZCP only & 90.94 $\pm$ 0.40 & 90.73 $\pm$ 0.37 & 72.48 $\pm$ 0.60 & 72.03 $\pm$ 0.57 & 46.61 $\pm$ 0.35 & 46.20 $\pm$ 0.42 & 6.89 & 8.00 \\
BOHAMIANN & 91.25 $\pm$ 0.14 & 90.97 $\pm$ 0.24 & 72.88 $\pm$ 0.35 & 72.36 $\pm$ 0.42 & 46.90 $\pm$ 0.27 & 46.44 $\pm$ 0.32 & 6.06 & 6.00 \\
RankNAS & 91.36 $\pm$ 0.16 & 91.06 $\pm$ 0.11 & 72.95 $\pm$ 0.26 & 72.30 $\pm$ 0.22 & 46.84 $\pm$ 0.16 & 46.42 $\pm$ 0.18 & 5.78 & 6.00 \\
BANANAS & 91.34 $\pm$ 0.18 & 91.00 $\pm$ 0.12 & 73.11 $\pm$ 0.19 & 72.46 $\pm$ 0.34 & 46.94 $\pm$ 0.32 & 46.44 $\pm$ 0.33 & 5.06 & 5.00 \\
ZCPBP Warmup & 91.28 $\pm$ 0.19 & 91.00 $\pm$ 0.24 & 73.00 $\pm$ 0.36 & 72.45 $\pm$ 0.50 & 46.79 $\pm$ 0.23 & 46.44 $\pm$ 0.32 & 4.50 & 5.00 \\
ReNAS & 91.40 $\pm$ 0.15 & 91.02 $\pm$ 0.20 & 73.14 $\pm$ 0.15 & 72.60 $\pm$ 0.27 & 46.91 $\pm$ 0.22 & \textbf{46.56 $\pm$ 0.24} & 3.00 & 3.00 \\
PWLNAS & \textbf{91.43 $\pm$ 0.12} & \textbf{91.13 $\pm$ 0.14} & \textbf{73.15 $\pm$ 0.14} & 72.59 $\pm$ 0.36 & \textbf{47.06 $\pm$ 0.24} & 46.53 $\pm$ 0.31 & 2.83 & 3.00 \\
\midrule
ppNAS (ours) & 91.34 $\pm$ 0.13 & 91.12 $\pm$ 0.14 & 73.09 $\pm$ 0.25 & \textbf{72.66 $\pm$ 0.34} & 46.88 $\pm$ 0.26 & 46.53 $\pm$ 0.29 & \textbf{2.28} & \textbf{2.00} \\
\bottomrule
\end{tabular}
}
\end{table}

\pagebreak
\section{Different search spaces}\label{sec:different_search_spaces}
We run our main experiments on the NB201 search spaces. We here repeat the main experiment on the NB101 search space. The results are in \Cref{tab:nb101_results}. Our proposed \framework again performs best.

\begin{table}[!htbp]
    \centering
    \caption{NB101 CIFAR10 search space.}
    \label{tab:nb101_results}
    \begin{tabular}{lll|ll}
\toprule
& \multicolumn{2}{c}{CIFAR10} & \multicolumn{2}{c}{Rank $\downarrow$} \\
Method &  Final Acc & AUC & Mean rank & Median rank \\
\midrule
Random search & 94.17 $\pm$ 0.23 & 94.03 $\pm$ 0.30 & 9.17 & 9.00 \\
ProxyBO & 94.26 $\pm$ 0.22 & 94.06 $\pm$ 0.21 & 8.33 & 8.00 \\
ZCP only & 94.20 $\pm$ 0.36 & 94.01 $\pm$ 0.34 & 7.50 & 9.00 \\
RankNAS & 94.42 $\pm$ 0.20 & 94.18 $\pm$ 0.15 & 6.67 & 7.00 \\
ZCPBP Warmup & 94.47 $\pm$ 0.18 & 94.15 $\pm$ 0.23 & 6.50 & 5.50 \\
ReNAS & 94.42 $\pm$ 0.13 & 94.19 $\pm$ 0.09 & 5.67 & 6.00 \\
BANANAS & 94.55 $\pm$ 0.15 & 94.23 $\pm$ 0.13 & 4.67 & 5.00 \\
PWLNAS & \textbf{94.56 $\pm$ 0.11} & 94.26 $\pm$ 0.15 & 3.17 & 3.00 \\
BOHAMIANN & 94.49 $\pm$ 0.17 & 94.26 $\pm$ 0.12 & 2.17 & 2.00 \\
\midrule
ppNAS (ours) & 94.54 $\pm$ 0.15 & \textbf{94.31 $\pm$ 0.17} & \textbf{1.17} & \textbf{1.00} \\
\bottomrule
\end{tabular}
\end{table}

\begin{table}[]
    \centering
        \caption{TNB autoencoder search space.}
    \label{tab:placeholder}
    \begin{tabular}{lllll}
\toprule
& \multicolumn{2}{c}{AutoEncoder} & \multicolumn{2}{c}{Rank $\downarrow$} \\
Method &  Final Acc & AUC & Mean rank & Median rank \\
\midrule
Random search & 0.73 $\pm$ 0.02 & 0.70 $\pm$ 0.02 & 9.60 & 10.00 \\
ReNAS & 0.74 $\pm$ 0.01 & 0.71 $\pm$ 0.02 & 8.00 & 8.00 \\
RankNAS & 0.73 $\pm$ 0.02 & 0.71 $\pm$ 0.03 & 7.00 & 7.00 \\
ZCPBP Warmup & 0.73 $\pm$ 0.01 & 0.71 $\pm$ 0.02 & 6.40 & 6.00 \\
PWLNAS & 0.74 $\pm$ 0.01 & 0.71 $\pm$ 0.02 & 6.00 & 6.00 \\
ZCP only & 0.73 $\pm$ 0.02 & 0.71 $\pm$ 0.03 & 5.60 & 5.00 \\
BANANAS & 0.74 $\pm$ 0.01 & 0.72 $\pm$ 0.02 & 4.20 & 4.00 \\
ProxyBO & 0.74 $\pm$ 0.01 & 0.72 $\pm$ 0.02 & 4.00 & 3.00 \\
BOHAMIANN & 0.74 $\pm$ 0.01 & 0.72 $\pm$ 0.01 & 2.60 & 2.00 \\
\midrule
ppNAS (ours) & \textbf{0.74 $\pm$ 0.01} & \textbf{0.72 $\pm$ 0.01} & \textbf{1.60} & \textbf{1.00} \\
\bottomrule
\end{tabular}

\end{table}

\pagebreak
\newpage
\section{Per-proxy results}\label{sec:per_proxy_results_extended}
\begin{figure}[htbp]
    \centering
    \includegraphics[width=\linewidth]{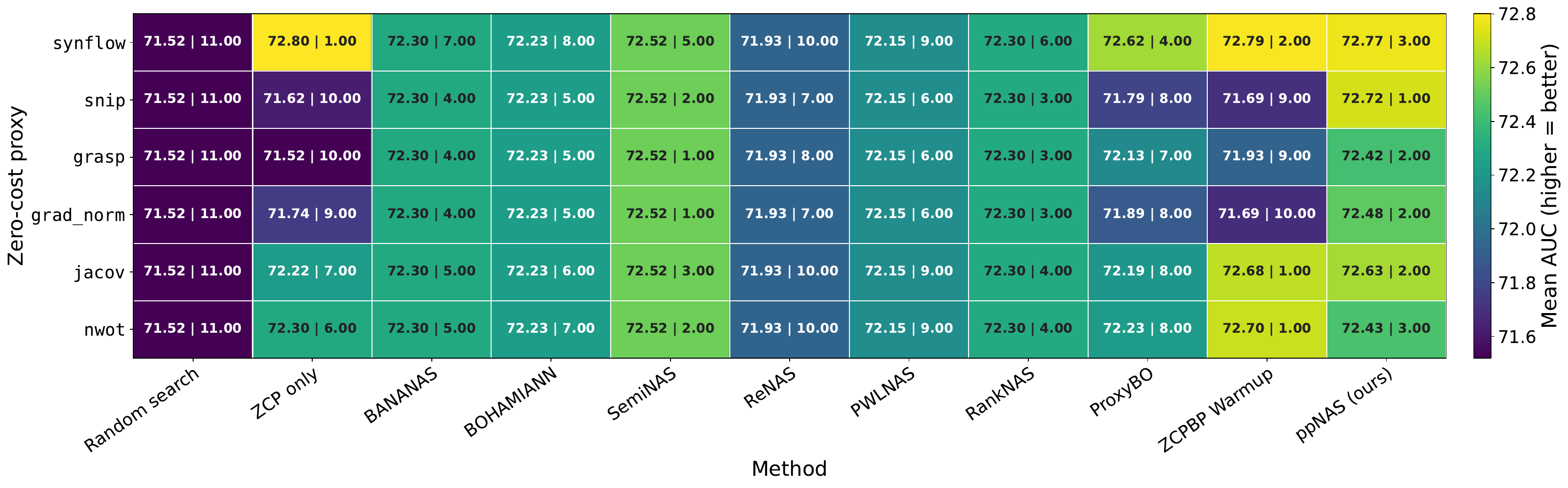}
    \caption{Per-proxy results for CIFAR100 from NB201.}
    \label{fig:heatmap_nb201_cifar100}
\end{figure}

\begin{figure}[htbp]
    \centering
    \includegraphics[width=\linewidth]{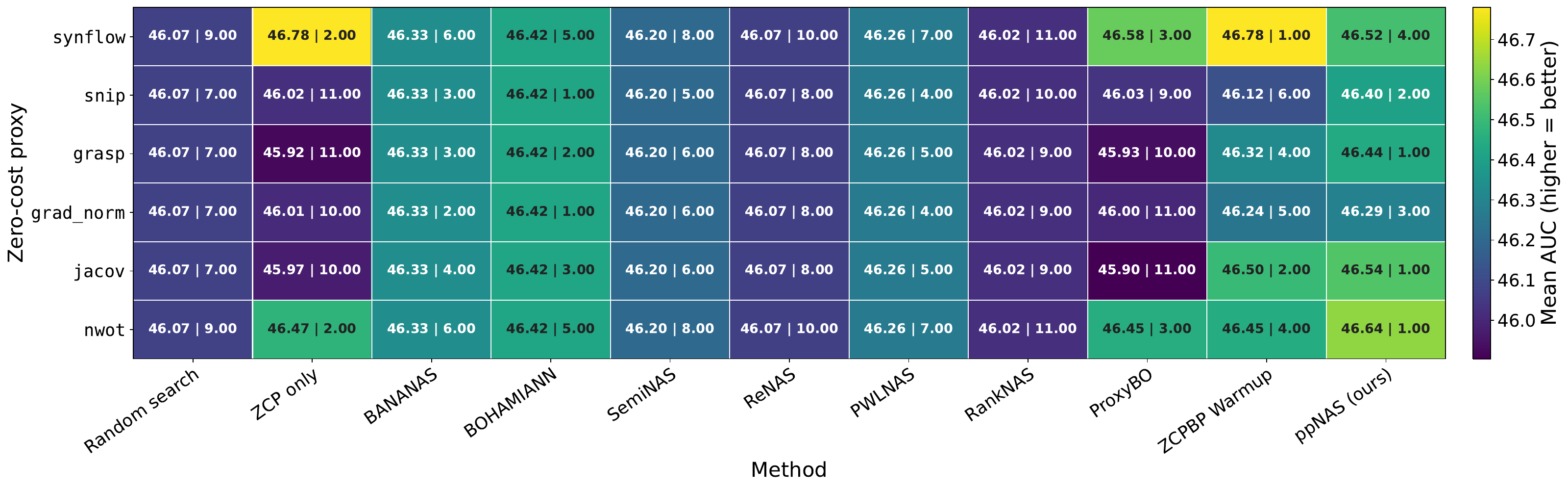}
\caption{Per-proxy results for ImageNet from NB201.}
    \label{fig:heatmap_nb201_imagenet}
\end{figure}

We visualize further per-proxy performances in \Cref{fig:heatmap_nb201_cifar100,fig:heatmap_nb201_imagenet}. Our observations are in line with the main results: \framework can use a wide range of ZCPs to construct ranking supervision from unsupervised data. It consistently is in the top-3 methods. 

\clearpage
\subsection{Stress-test to inverted proxy correlations}
\begin{figure}
    \centering
    \includegraphics[width=\linewidth]{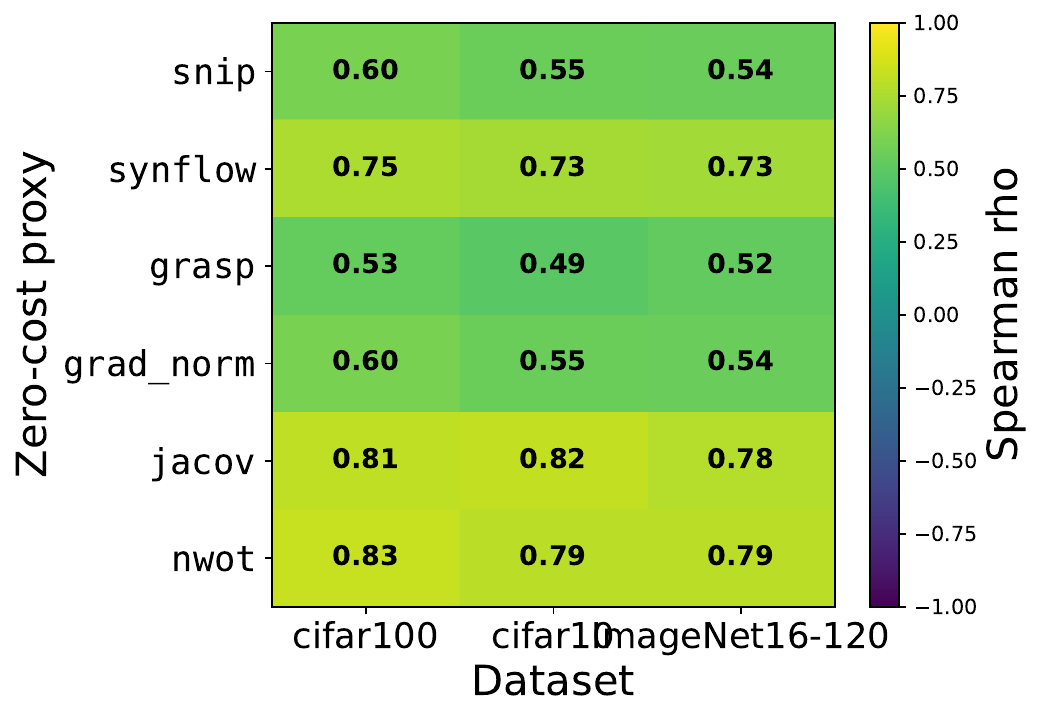}
    \caption{Spearman ranking correlation between zero-cost proxies and validation accuracy for CIFAR10 from the NB101 benchmark.}
    \label{fig:nb201_proxy_correlation}
\end{figure}

\begin{figure}
    \centering
    \includegraphics[width=0.5\linewidth]{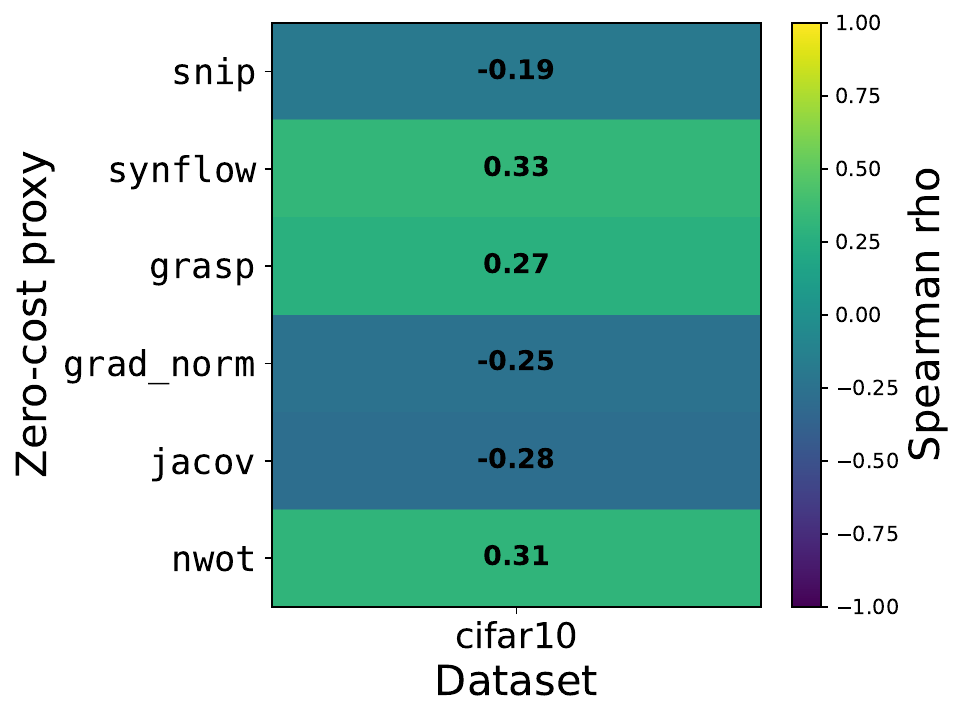}
    \caption{Spearman ranking correlation between zero-cost proxies and validation accuracy for CIFAR10 from the NB101 benchmark.}
    \label{fig:nb101_proxy_correlation}
\end{figure}

\begin{figure}
    \centering
    \includegraphics[width=\linewidth]{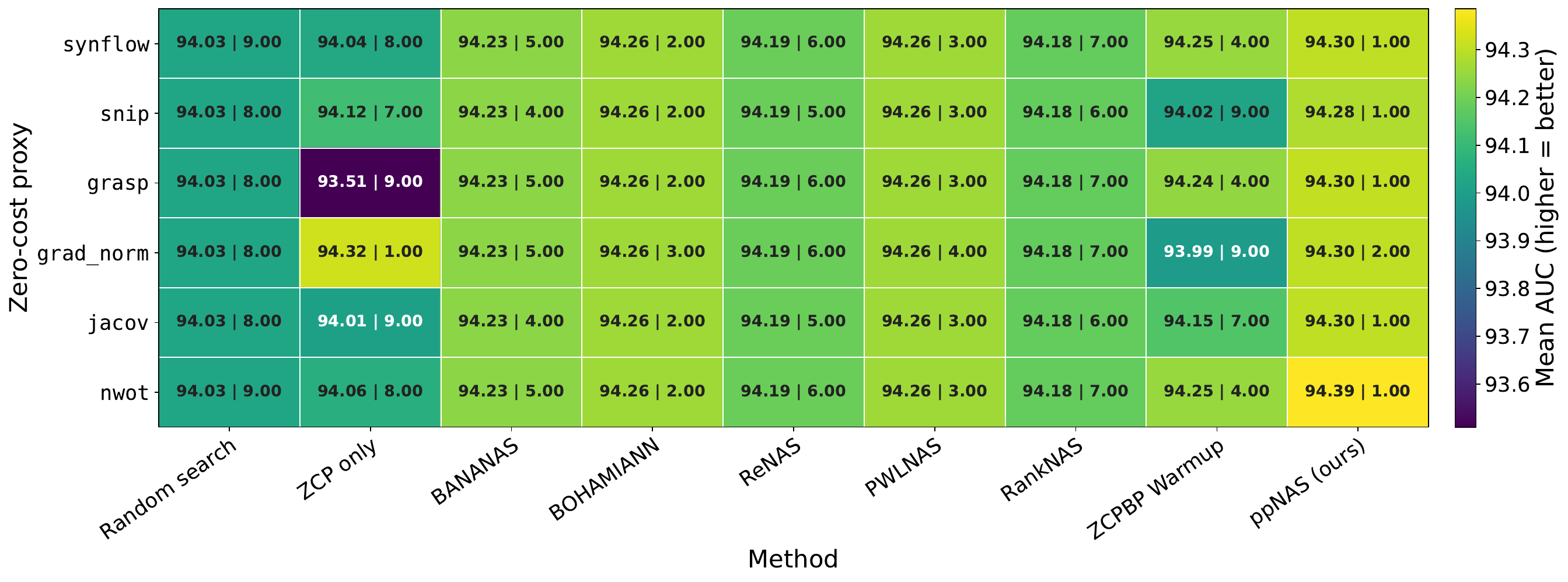}
    \caption{Per-proxy results for CIFAR10 from NB101}
    \label{fig:nb101_heatmap}
\end{figure}

\clearpage
\section{Runtime analysis}
\begin{figure}[htbp]
    \centering
    \includegraphics[width=\linewidth]{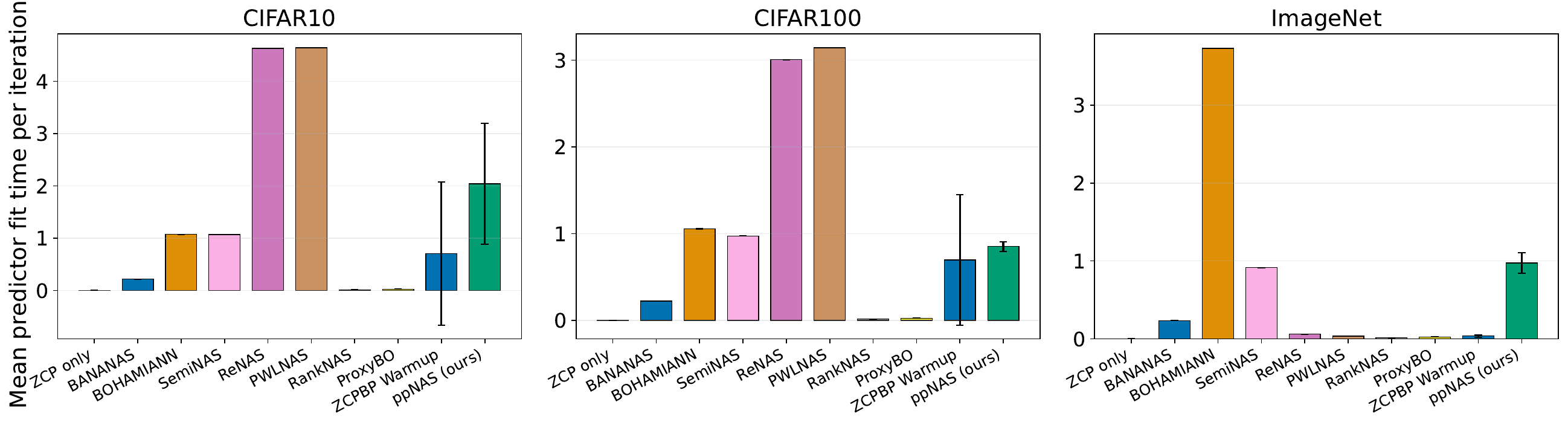}
    \caption{Runtimes. \framework is comparable to the baselines.}
    \label{fig:runtimes}
\end{figure}

We visualize the average runtimes across the NAS iterations in \Cref{fig:runtimes}. Runtimes of \framework are in line with the baselines.

\end{document}